\documentclass[preprint,12pt]{elsarticle}

\usepackage{amssymb}
\usepackage{amsmath}
\usepackage{graphicx}
\usepackage{booktabs}
\usepackage{multirow}
\usepackage{tabularx}
\usepackage{hyperref} 

\journal{Pattern Recognition}

\begin{document}

\begin{frontmatter}

%% Title, authors and addresses

%% use the tnoteref command within \title for footnotes;
%% use the tnotetext command for theassociated footnote;
%% use the fnref command within \author or \affiliation for footnotes;
%% use the fntext command for theassociated footnote;
%% use the corref command within \author for corresponding author footnotes;
%% use the cortext command for theassociated footnote;
%% use the ead command for the email address,
%% and the form \ead[url] for the home page:
%% \title{Title\tnoteref{label1}}
%% \tnotetext[label1]{}
%% \author{Name\corref{cor1}\fnref{label2}}
%% \ead{email address}
%% \ead[url]{home page}
%% \fntext[label2]{}
%% \cortext[cor1]{}
%% \affiliation{organization={},
%%             addressline={},
%%             city={},
%%             postcode={},
%%             state={},
%%             country={}}
%% \fntext[label3]{}

\title{A Comprehensive Review of 3D Object Detection in Autonomous Driving: Technological Advances and Future Directions}

%% use optional labels to link authors explicitly to addresses:
\author[label1]{Yu Wang}
\author[label1]{Shaohua Wang}
\author[label1]{Yicheng Li}
\author[label2]{Mingchun Liu}

\affiliation[label1]{organization={Institute of Automotive Engineering, Jiangsu University},%Department and Organization
            addressline={No. 301 Xuefu Road}, 
            city={Zhenjiang},
            postcode={212013}, 
            state={Jiangsu},
            country={China}}

\affiliation[label2]{organization={Higer Bus Company Limited, Institute of Advanced Technology},
            addressline={288 Suhong East Road, Suzhou Industrial Park},
            city={Suzhou},
            postcode={215127},
            state={Jiangsu},
            country={China}}

%% Abstract
\begin{abstract}
%% Text of abstract
In recent years, 3D object perception has become a crucial component in the development of autonomous driving systems, providing essential environmental awareness. However, as perception tasks in autonomous driving evolve, their variants have increased, leading to diverse insights from industry and academia. Currently, there is a lack of comprehensive surveys that collect and summarize these perception tasks and their developments from a broader perspective. This review extensively summarizes traditional 3D object detection methods, focusing on camera-based, LiDAR-based, and fusion detection techniques. We provide a comprehensive analysis of the strengths and limitations of each approach, highlighting advancements in accuracy and robustness. Furthermore, we discuss future directions, including methods to improve accuracy such as temporal perception, occupancy grids, and end-to-end learning frameworks. We also explore cooperative perception methods that extend the perception range through collaborative communication. By providing a holistic view of the current state and future developments in 3D object perception, we aim to offer a more comprehensive understanding of perception tasks for autonomous driving. Additionally, we have established an active repository to provide continuous updates on the latest advancements in this field, accessible at: \href{https://github.com/Fishsoup0/Autonomous-Driving-Perception}{https://github.com/Fishsoup0/Autonomous-Driving-Perception}.
\end{abstract}

%%Graphical abstract
\begin{graphicalabstract}
\includegraphics[width=1\linewidth]{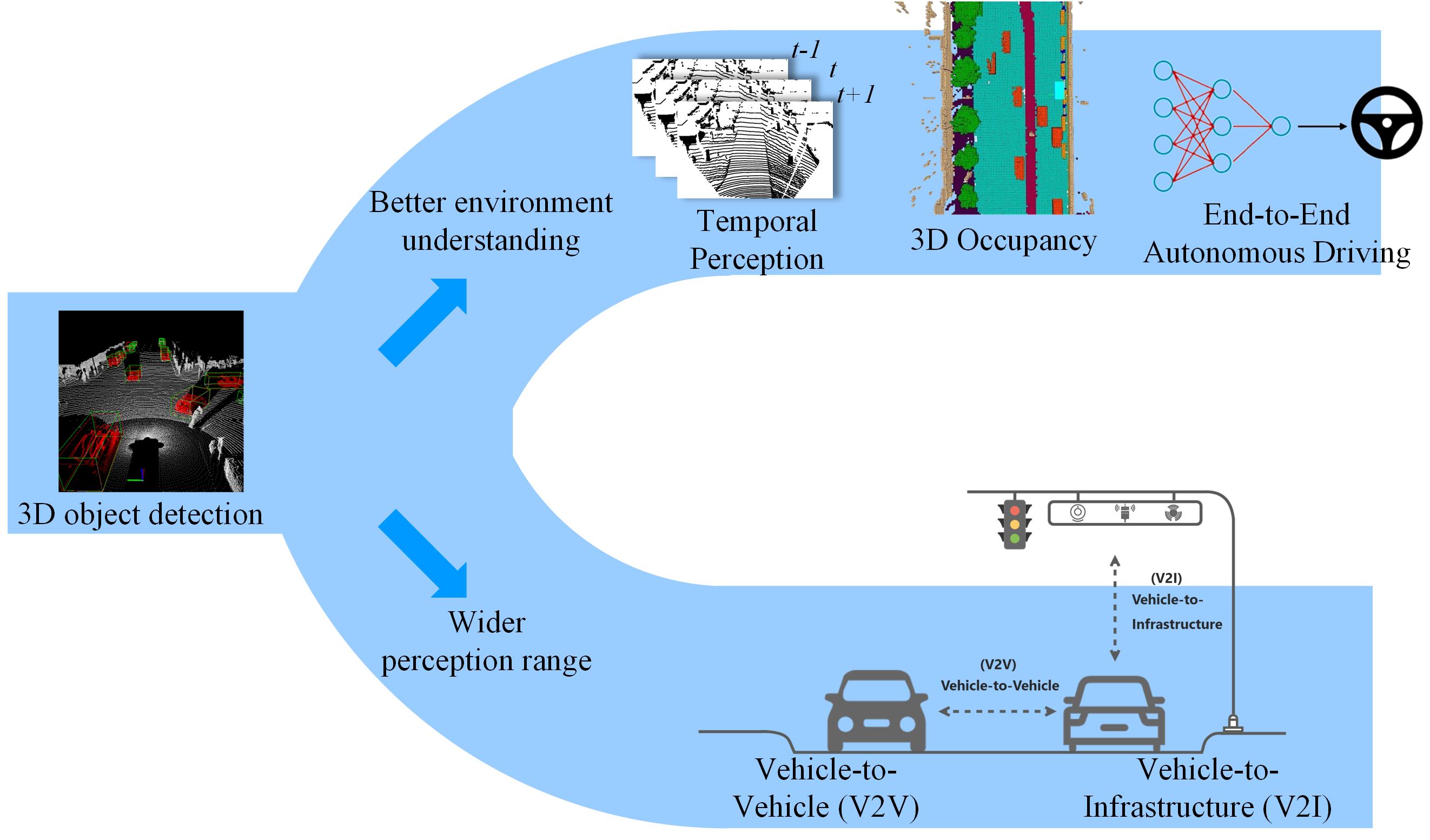}
\end{graphicalabstract}

%%Research highlights
\begin{highlights}
\item Comprehensive analysis of 3D object detection methods: camera-based, LiDAR, and fusion.
\item Provides a holistic view of the evolution and future trends in 3D object perception.
\item Provides a holistic view of 3D perception in autonomous driving environments to facilitate research insights.
\end{highlights}

%% Keywords
\begin{keyword}
%% keywords here, in the form: keyword \sep keyword

%% PACS codes here, in the form: \PACS code \sep code
3D Object Detection \sep Autonomous Driving \sep Deep learning \sep Computer vision
%% MSC codes here, in the form: \MSC code \sep code
%% or \MSC[2008] code \sep code (2000 is the default)

\end{keyword}

\end{frontmatter}

%% Add \usepackage{lineno} before \begin{document} and uncomment 
%% following line to enable line numbers
%% \linenumbers

%% main text
%%

%% Use \section commands to start a section
\section{Introduction}
\label{sec1}
%% Labels are used to cross-reference an item using \ref command.

The rapid development of autonomous driving technology aims to revolutionize transportation by enhancing safety, efficiency, and convenience. Among the capabilities of autonomous vehicles, 3D object detection is key to achieving precise environmental perception. 3D object detection plays a crucial role in identifying and understanding objects around the vehicle, facilitating tasks such as object tracking, obstacle avoidance, and path planning. This technology utilizes various sensors, including cameras, LiDAR , and multi-sensor fusion, each with its own characteristics and advantages, making them suitable for different scenarios.

As deep learning continues to advance and computing power increases, achieving the goal of autonomous driving using deep learning technologies seems within reach. Both academia and industry have invested significant resources in research, leading to the rapid development of autonomous driving technology and the emergence of new techniques and methods\cite{yurtsever2020survey,kiran2021deep,zhao2023autonomous}. Some of these methods refine 3D object perception to achieve higher granularity in detection, while others focus on achieving broader detection ranges. Some even go as far as to directly bypass the perception module in favor of end-to-end driving. Therefore, it is necessary to summarize and analyze the latest 3D object detection methods and compare them with different development trends to gain a comprehensive understanding of the entire field of autonomous driving perception.

To achieve this goal, we have comprehensively reviewed the latest 3D object detection methods, including camera-based, LiDAR-based, and fusion-based approaches. We have also summarized new directions in 3D object detection, including temporal perception, 3D occupancy grids, end-to-end autonomous driving, and collaborative perception. Additionally, we have compiled and summarized the datasets and evaluation metrics used by different methods to better facilitate research comparisons. Compared to previous reviews that focused solely on summarizing image-based\cite{ma20233d}, LiDAR-based\cite{wu2020deep,fernandes2021point}, or multimodal approaches\cite{huang2022multi,qian20223d,mao20233d,arnold2019survey}, we provide a comprehensive and broad perspective on autonomous driving perception, accompanied by comparative analysis, offering readers a fresh viewpoint.

The main contributions of this paper are as follows:

1. To our knowledge, this is the first time that different development trends in autonomous driving environmental perception have been summarized and analyzed, providing a holistic view of the evolution and future trends in 3D object perception.

2. We provide a comprehensive summary, classification, and analysis of the latest methods in camera-based, LiDAR-based, and fusion-based 3D object detection.

3. We offer a panoramic view of perception in autonomous driving environments, not only summarizing the perception methods comprehensively but also compiling datasets and evaluation metrics used by different methods to promote research insights.

\begin{figure}[t]%% placement specifier
%% Use \includegraphics command to insert graphic files. Place graphics files in 
%% working directory.
\centering%% For centre alignment of image.
\includegraphics[width=1\linewidth]{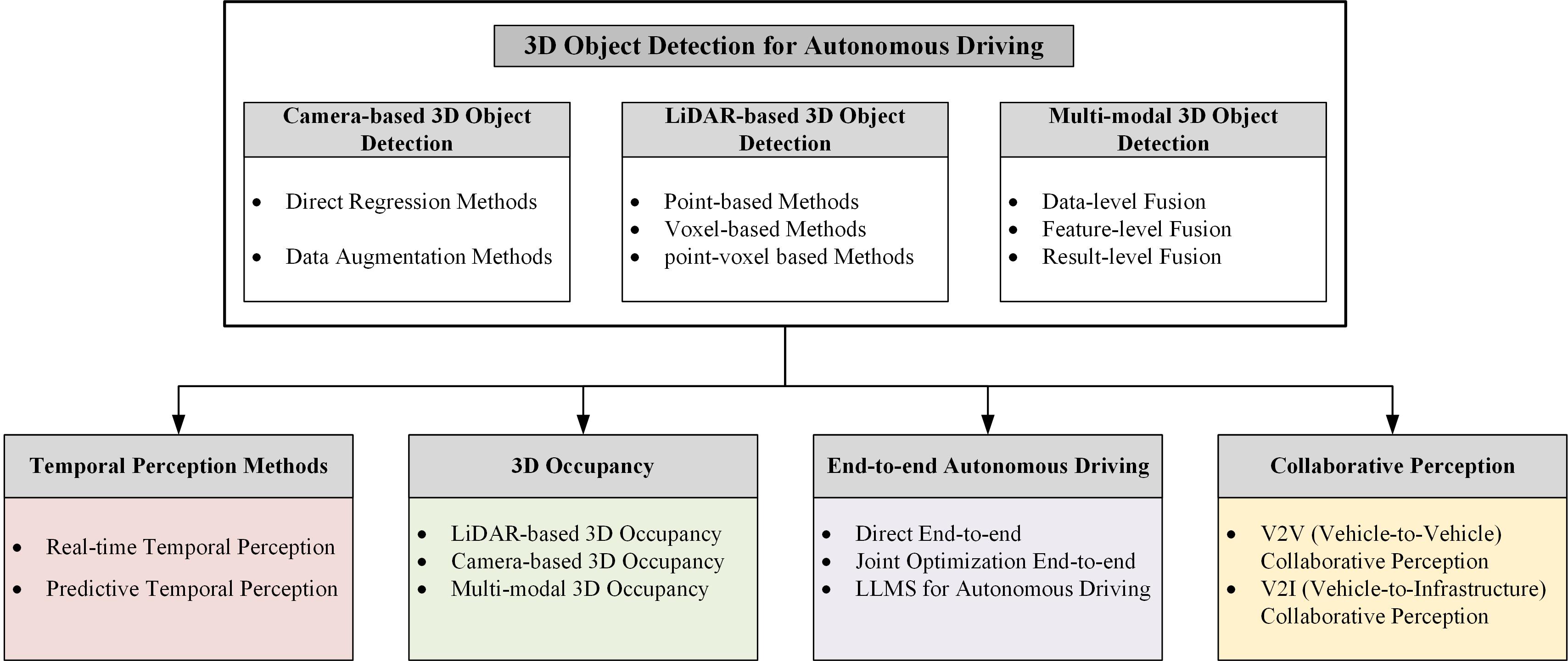}
%% Use \caption command for figure caption and label.
\caption{Chapter Structure and Classification of Autonomous Driving Perception Techniques.}\label{fig1}
%% https://en.wikibooks.org/wiki/LaTeX/Importing_Graphics#Importing_external_graphics
\end{figure}

The structure of this paper is as follows.Figure\ref{fig1} provides an overview of the chapter structure, outlining the key sections and their relationships. In Section 2, we introduce the datasets and evaluation metrics for 3D object detection in autonomous driving and its development trends. In Section 3, we provide a comprehensive summary of camera-based, LiDAR-based, and fusion-based methods for single-vehicle perception. In Section 4, we summarize the development trends in 3D object detection for autonomous driving, including temporal perception, 3D occupancy grids, end-to-end autonomous driving, and V2X collaborative perception. In Section 5, we discuss and summarize the technologies in autonomous driving environmental perception and propose future research directions. Finally, in Section 6, we conclude the paper.

\section{Background}
\subsection{Problem Definition}

In the field of autonomous driving, 3D perception technology is crucial for enabling vehicles to have a comprehensive understanding of their surrounding environment. This paper involves various 3D perception tasks and related tasks, such as 3D object detection, temporal perception, 3D occupancy grids, end-to-end autonomous driving perception, and collaborative perception. Each task presents its specific challenges and requirements.

3D Object Detection:
The task of 3D object detection aims to identify and locate various objects (such as vehicles, pedestrians, and obstacles) from sensor data, accurately determining their position, size, orientation, and class in three-dimensional space. Systems use LiDAR or cameras to capture environmental data and apply algorithms to identify different objects. To simplify the task, a bounding box is usually used to represent detected objects. Each bounding box includes the central point location, dimensions (length, width, height), the orientation angle of the object, and the object class.

\begin{equation}
B = [x, y, z, l, w, h, \theta, \text{class}]
\label{eq:bounding_box}
\end{equation}

\textbf{Temporal Perception:}
Temporal perception tasks focus on detecting and tracking objects in dynamic scenes, requiring the system to process continuous sensor data streams in real-time while maintaining real-time object detection. The system must quickly respond to environmental changes, providing stable and reliable environmental information. Temporal perception is mainly divided into real-time temporal perception and predictive temporal perception.

\textbf{3D Occupancy Grid: }
Existing 3D bounding box methods have limitations in capturing the geometric details of objects and handling undefined classes. To more comprehensively simulate the geometry and semantics of 3D objects, the task of 3D occupancy prediction is proposed. This task aims to understand the state of each voxel in 3D space, estimating its occupancy status and semantic label. This approach significantly enhances the accuracy of path planning and collision detection by refining spatial representation. The goal of this task is to understand the state of each voxel in 3D space and estimate its occupancy status and semantic label.

\textbf{End-to-End Autonomous Driving :}
This method directly takes sensor data as input and produces driving decisions as output, without requiring manually defined feature extraction and classification steps, although some methods involve joint optimization. It simplifies the system architecture but requires the overall performance to be highly optimized to ensure that every processing step directly influences the final driving decision. This method is typically evaluated directly on autonomous driving outcomes, such as decision accuracy and collision rate.

\textbf{Collaborative Perception:}
Collaborative perception enhances a single vehicle's perception capabilities by sharing information between vehicles (V2V) or between vehicles and infrastructure (V2I). By integrating information, the system improves perception accuracy and robustness in complex environments. The final goal of detection remains 3D object perception, but since it uses more sensors, the target design scope is broader. Additionally, due to steps like data transmission and feature matching, researchers also study the robustness of algorithms, including factors such as delay errors, localization errors, and data noise.

\subsection{Evaluation Metrics}

In the field of 3D object detection, evaluation metrics are primarily used to measure the accuracy of algorithms in object recognition. The most intuitive metric is the 3D Intersection over Union (3D IoU), which measures the overlap between the predicted 3D bounding box and the ground truth 3D bounding box. The formula for 3D IoU is:

\begin{equation}
   \text{3D IoU} = \frac{\text{Area of Intersection}}{\text{Area of Union}} 
\end{equation}

To qualitatively assess whether the predicted bounding box accurately detects the object, an IoU threshold (e.g., 0.5 or 0.7) is usually set. Based on this threshold, each ground truth box is assigned the detection result with the maximum IoU, and detection results can be classified as True Positives (TP), False Positives (FP), or False Negatives (FN). Further, precision and recall metrics can be introduced to evaluate performance:
\begin{equation}
    \text{Precision} = \frac{\text{TP}}{\text{TP} + \text{FP}}
\end{equation}

\begin{equation}
\text{Recall} = \frac{\text{TP}}{\text{TP} + \text{FN}}
\end{equation}

Considering the impact of confidence on detection results, predicted results can be sorted by confidence from high to low. By selecting a series of confidence thresholds, we can obtain a series of different precision and recall combinations. These data points can be plotted to form a Precision-Recall (PR) curve. The area under the PR curve (AP, Average Precision) is an important metric for evaluating overall detection performance, usually estimated by numerically integrating the PR curve:

\begin{equation}
\text{AP} = \int_{0}^{1} P(r) \, dr
\end{equation}

where \(P(r)\) represents the precision at recall \(r\). Additionally, KITTI\cite{geiger2012we} introduces AP\(_\text{bev}\), which evaluates IoU from a bird's-eye view without considering height.NuScenes\cite{caesar2020nuscenes} introduces the AP\(_\text{center}\) metric, where predicted objects are matched with ground truth based on their center positions being within a specified distance. This approach also accounts for size and orientation errors, leading to the development of the NuScenes Detection Score (NDS). Waymo\cite{sun2020scalability} introduces the AP\(_\text{hungarian}\) algorithm, which utilizes the Hungarian algorithm to match ground truth with predictions, and proposes a heading-weighted AP (APH) metric.

In the evaluation of end-to-end autonomous driving systems, open-loop systems and closed-loop systems are considered\cite{li2024ego,zhai2023rethinking,chen2024end}. Open-loop evaluation focuses on the system's prediction capabilities and object detection accuracy, using metrics such as trajectory deviation and prediction error to measure the system's response to pre-recorded data. Closed-loop evaluation, however, simulates real driving scenarios to assess the system's overall performance, focusing on collision rate, lane-keeping rate, driving comfort, and other indicators to comprehensively evaluate the system's safety and operational stability.

\subsection{Datasets}
Against the backdrop of the rapid development of autonomous driving and vehicular communication technologies, the role of high-quality datasets in driving technological progress has become increasingly important. Pioneering datasets such as KITTI, NuScenes, and Waymo Open Dataset have provided a variety of real-world scenarios for research in the field of single-vehicle perception by offering LiDAR point clouds and 360-degree panoramic camera data, greatly promoting the development of autonomous driving technology. As illustrated in Table \ref{tab:datasets}, these datasets have laid a solid foundation for the subsequent construction of end-to-end autonomous driving and 3D occupancy grid datasets.

In the research of end-to-end autonomous driving, open-loop testing usually relies on datasets such as NuScenes, while closed-loop testing relies more on simulators like Carla\cite{dosovitskiy2017carla}, LGSVL Simulator\cite{rong2020lgsvl}, and Microsoft AirSim\cite{shah2018airsim}. These tools provide controllable virtual environments to simulate complex traffic conditions. 3D occupancy grid datasets have also been widely applied and developed on this basis. Datasets like Semantic KITTI\cite{behley2019semantickitti} and KITTI-360\cite{liao2022kitti} provide a more refined spatial representation by dividing the environment into small grid cells and determining the occupancy status of each cell, greatly enhancing the accuracy of path planning and collision detection. Datasets such as Occ3D\cite{tian2024occ3d} and Lyft Level 5\cite{li2023large} further promote the development of 3D occupancy grid technology by integrating data from multiple sensors.

With the deepening of research, collaborative perception in the fields of Vehicle-to-Vehicle (V2V), Vehicle-to-Infrastructure (V2I), and Infrastructure-to-Infrastructure (I2I) communication has become a new research hotspot. This has driven the development of V2X communication systems and spawned several key datasets. OPV2V\cite{xu2022opv2v}, as the first simulation dataset focusing on V2V collaborative perception, has laid the foundation for this field. V2XSet\cite{xu2022v2x} and V2X-Sim\cite{li2022v2x} further expand the research on various V2X scenarios, covering more complex collaborative perception scenarios. However, the difference between simulation datasets and the real world has prompted researchers to develop datasets closer to actual application scenarios. The emergence of large-scale real-world datasets such as V2X-Real\cite{xiang2024v2x} and DAIR-V2X\cite{yu2022dair} has promoted the development of collaborative perception research. The TUMTraf-V2X\cite{zimmer2024tumtraf} dataset further promotes the application research in complex multi-agent environments by collecting data at complex intersections and providing 3D annotations with precise GPS and IMU data.

The development of datasets will continue to advance in several directions. First, the integration of multimodal data will be further enhanced, especially in complex environments. Second, combining real-world and simulation data to narrow the gap between the two will be a key area to improve the reliability of algorithms in practical applications. In addition, increasing the collection of large-scale long-term data, covering different weather, lighting conditions, and traffic flows, will help improve the model's generalization ability in various complex environments.

\begin{table}[ht]
\centering
\caption{Overview of Datasets for Different 3D Perception Tasks (C: Camera, L: LiDAR)}
\resizebox{\textwidth}{!}{
\begin{tabular}{|l|l|l|l|l|l|l|l|}
\hline
\textbf{Task Type} & \textbf{Year} & \textbf{Dataset Name} & \textbf{Environment} & \textbf{Sensor Type } & \textbf{Frames} & \textbf{Unique Features} & \textbf{Other Tasks} \\ \hline

\multirow{3}{*}{3D Object Detection} & 2012 & KITTI\cite{geiger2012we} & Real & L, C & 15k & Pioneering dataset, widely used & 2D object detection, stereo vision \\ \cline{2-8}
                           & 2019 & NuScenes\cite{caesar2020nuscenes} & Real & L, C, Radar & 400k  & 360-degree coverage, diverse weather & Object tracking, sensor fusion \\ \cline{2-8}
                           & 2020 & Waymo Open\cite{sun2020scalability} & Real & L, C & 200k  & Large-scale data, diverse environments & Tracking, segmentation \\ \hline

\multirow{4}{*}{3D Occupancy} & 2019 & Semantic KITTI\cite{behley2019semantickitti} & Real & L & 43k  & Semantic segmentation support & SLAM \\ \cline{2-8} 
                           & 2019 & KITTI-360\cite{liao2022kitti} & Real & L, C & 27k  & 360-degree scenes, extended KITTI & Mapping, localization \\ \cline{2-8} 
                           & 2020 & Occ3D\cite{tian2024occ3d} & Real & L, C & 120k  & Comprehensive 3D occupancy  & Semantic segmentation \\ \cline{2-8}
                           & 2020 & Lyft Level 5\cite{li2023large} & Real & L, C & 55k  & high-resolution sensor data & Autonomous driving, localization \\ \hline

\multirow{7}{*}{Collaborative Perception} & 2021 & OPV2V\cite{xu2022opv2v} & Simulated & L, C & 12k & First V2V dataset & V2V communication \\ \cline{2-8} 
                           & 2021 & V2X-Sim\cite{li2022v2x} & Simulated & L, C, GPS/IMU & 11,464  & Multi-modal support & V2V, V2I communication \\ \cline{2-8}
                           & 2022 & V2XSet\cite{xu2022v2x} & Simulated & L, C & 12k & Extended to V2X & V2V, V2I communication \\ \cline{2-8}
                           & 2022 & Rope3D\cite{ye2022rope3d} & Real & C, GPS/IMU & 50,009  & Focus on camera collaboration & Camera-based localization \\ \cline{2-8}
                           & 2022 & DAIR-V2X\cite{yu2022dair} & Real & L, C, GPS/IMU & 38,845  & All collaboration modes, dense urban & V2I, sensor fusion \\ \cline{2-8}
                           & 2024 & V2X-Real\cite{xiang2024v2x} & Real & L, C, GPS/IMU & 33k   & All collaboration modes, dense urban & V2V, V2I, I2I communication \\ \cline{2-8}
                           & 2024 & TUMTraf-V2X\cite{zimmer2024tumtraf} & Real & L, C,Maps & 29,380  & All collaboration modes, dense urban & V2I, high-definition mapping \\ \hline

\end{tabular}
}
\label{tab:datasets}
\end{table}

\section{Current State of Perception Technology for Autonomous Vehicles}
In the field of autonomous driving, 3D object detection is crucial for achieving precise environmental perception by vehicles. This technology primarily relies on three core sensors: monocular cameras, stereo cameras, and LiDAR. Monocular cameras, known for their cost-effectiveness and widespread application, capture 2D images to provide visual information and are sensitive to color information, such as traffic lights and signs, which rely on camera recognition. However, they have limitations in acquiring depth information and are sensitive to environmental conditions, with performance significantly dropping at night or during rainy weather. In contrast, LiDAR constructs a 3D model of the environment by emitting lasers and measuring their reflection times, providing high-precision depth information, albeit with higher costs and data processing demands. As accuracy requirements continue to rise, multi-sensor fusion technology has emerged to integrate data from different sensors, aiming to compensate for the shortcomings of individual sensors and enhance detection accuracy and system robustness\cite{yeong2021sensor,xiang2023multi,kocic2018sensors}. However, this process involves complex steps such as sensor calibration, data alignment, and feature fusion, requiring significant computational resources. To provide an in-depth understanding, this paper highlights a series of representative and pioneering research achievements, particularly those with open-source code, enabling readers to more easily reproduce and understand the details and innovations of these technologies.

\subsection{Camera-Based 3D Object Detection}
Monocular camera-based 3D object detection infers the three-dimensional structure and position of objects within a scene by analyzing a single two-dimensional image. This task relies on recovering depth information from the 2D image, which is generated through a perspective projection process. In this process, a point P = (X, Y, Z) in the 3D world is mapped to a point p = (x, y) on the 2D image plane via the camera, following the pinhole camera model. Despite the challenges posed by scale ambiguity and the loss of depth information, deep learning techniques, particularly convolutional neural networks (CNNs), can effectively estimate depth from a single image by learning the relationship between 2D images and corresponding depth annotations. These techniques leverage visual cues within the image, such as texture gradients, edges, occlusions, and perspective diminishment, to reconstruct the 3D structure of the scene without the need for additional depth sensors. This demonstrates the powerful capability of neural networks in monocular vision systems to recover 3D information from 2D images.
\subsubsection{Direct Regression Methods}

\begin{figure}[t]%% placement specifier
%% Use \includegraphics command to insert graphic files. Place graphics files in 
%% working directory.
\centering%% For centre alignment of image.
\includegraphics[width=0.8\linewidth]{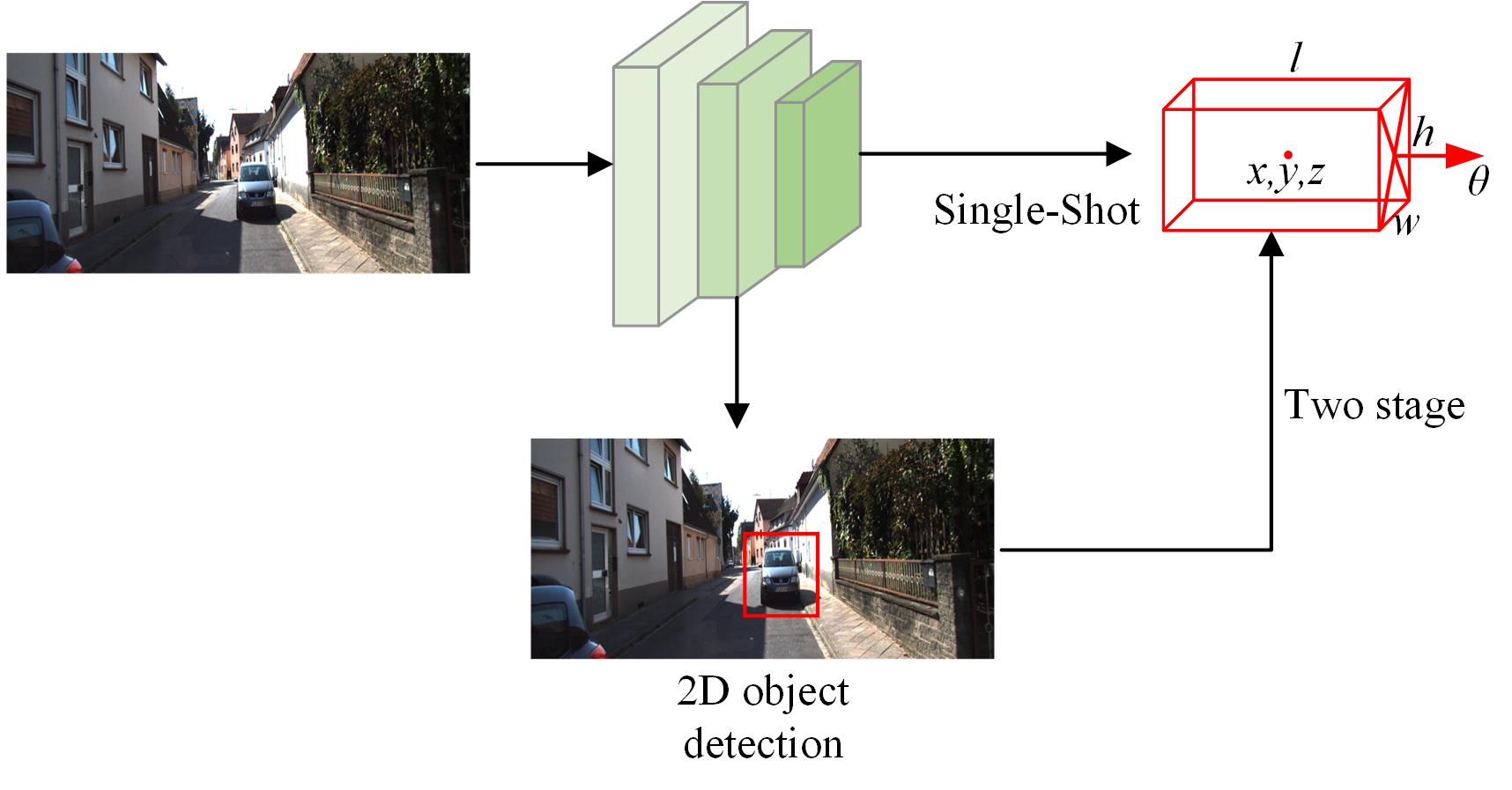}
%% Use \caption command for figure caption and label.
\caption{An illustration of direct regression  methods.}\label{fig2}
%% https://en.wikibooks.org/wiki/LaTeX/Importing_Graphics#Importing_external_graphics
\end{figure}

Direct regression methods primarily predict 3D object parameters directly from monocular images using end-to-end neural networks. These methods are typically simple and efficient, closely resembling the pattern of 2D object detection. As shown in Figure \ref{fig2}, direct regression methods can be categorized into single-stage and two-stage detection, each with its own approach to balancing speed and accuracy.

Single-stage detection methods extract features directly and infer spatial positions from these features. For example, YOLO3D\cite{yolo3d} extends the well-known YOLO\cite{redmon2018yolov3} algorithm framework from 2D object detection, directly predicting 3D bounding boxes from monocular images. The SMOKE\cite{liu2020smoke} algorithm parameterizes the representation of 3D bounding boxes, predicting the projection points of the 3D box center on the 2D image, and then uses these points to regress the object's size, position, and orientation. Subsequent methods like MonoDEL\cite{ma2021delving} and MonoCon\cite{liu2022monocon} further improve the accuracy of single-stage detection. FCOS3D\cite{wang2021fcos3d} is a 3D object detection algorithm based on monocular images that inherits and develops the FCOS algorithm, achieving accurate 3D object detection by decoupling 2D and 3D features.

Two-stage detection typically divides the detection process into two steps: first, regressing 2D bounding box information, and then extracting features in the ROI region to generate 3D box information. GUPNet\cite{lu2021geometry} employs CenterNet-based 2D detection and ROI feature extraction, with its core component being the GUP module. This module not only outputs depth values but also the uncertainty of these depth values, providing high-confidence depth estimates combined with a basic 3D detection head to achieve 3D object detection. MonoDis\cite{simonelli2019disentangling} detects 2D bounding boxes and corresponding features as candidate regions using a 2D object detection algorithm, instead of directly regressing the eight vertices of the 3D bounding box. It proposes a decoupled regression loss by setting a 10-tuple representation to replace the previous method, which regressed center, size, and rotation simultaneously, addressing the issue of different loss magnitudes during training.

\subsubsection{Based on Data Augmentation Methods}
\begin{figure}[t]%% placement specifier
%% Use \includegraphics command to insert graphic files. Place graphics files in 
%% working directory.
\centering%% For centre alignment of image.
\includegraphics[width=0.8\linewidth]{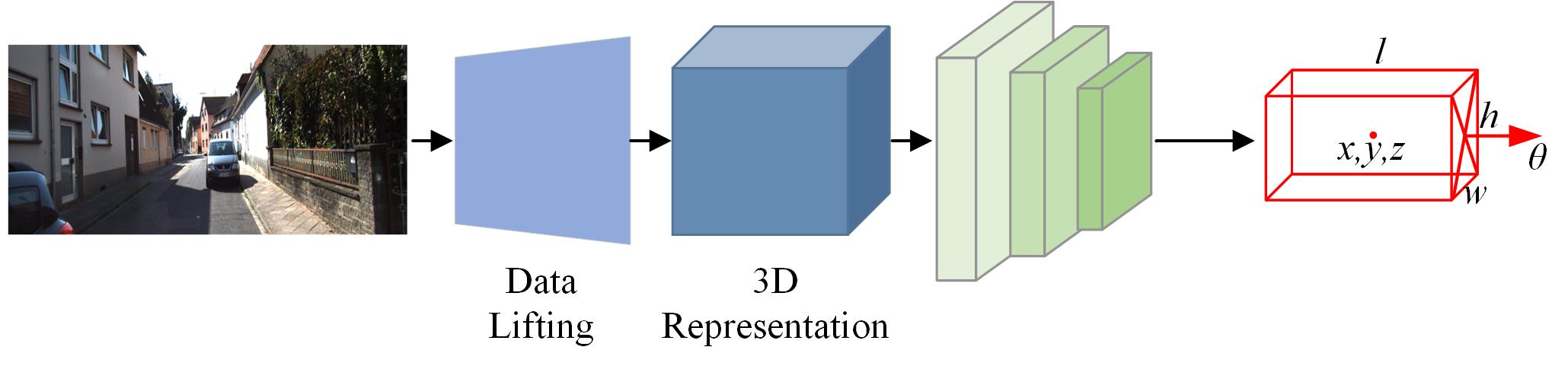}
%% Use \caption command for figure caption and label.
\caption{An illustration of methods based on data augmentation.}\label{fig3}
%% https://en.wikibooks.org/wiki/LaTeX/Importing_Graphics#Importing_external_graphics
\end{figure}
In the field of monocular 3D object detection, data augmentation methods have garnered extensive attention from researchers due to their potential to enhance detection performance. Directly predicting 3D objects from 2D images is inherently ill-posed, and errors in depth prediction are often the main factor leading to differences in performance between image and point cloud 3D detection. As a result, some methods focus on data augmentation rather than direct prediction. These methods typically employ monocular depth estimation models to generate pixel-level depth maps or depth features, integrating these depth details with monocular input detectors to improve detection accuracy. As shown in Figure \ref{fig3}, this approach usually necessitates additional depth network processing, which increases data volume and may impact real-time performance. Moreover, the accuracy of the depth estimation model is crucial, as its prediction errors can be propagated into the 3D object detection model, thus affecting the final detection performance. Moreover, the accuracy of the depth estimation model is crucial, as its prediction errors can be propagated into the 3D object detection model, thus affecting the final detection performance. For instance, the Pseudo-LiDAR\cite{wang2019pseudo} algorithm uses monocular depth estimation algorithms such as deep ordinal regression network(DORN) to generate depth maps, which are then reprojected into 3D space to form pseudo-LiDAR data. Subsequently, point cloud-based detection algorithms are utilized for 3D bounding box detection. The MF3D\cite{tung2017mf3d}algorithm adopts a different strategy by generating depth maps through a sub-network and combining the Region of Interest (RoI) with the depth map to regress the 3D positional information of the object. This approach avoids the need for two separate networks, enabling end-to-end training, thereby simplifying the process and improving efficiency.Recently, refining depth cues has also provided valuable insights. MonoCD\cite{yan2024monocd} enhances the accuracy and robustness of detection by utilizing the complementarity of global depth cues and geometric relationships without improving the precision of individual detection branches. OPEN\cite{hou2024openobjectwisepositionembedding}  through Object-wise Position Embedding, effectively injects object depth information, thereby enhancing the accuracy of 3D detection.

Benefiting from the advancements of attention mechanisms in non-local encoding and object detection, Transformer-based monocular 3D detectors have enhanced the global perception capability of models. MonoDTR\cite{huang2022monodtr} injects global depth information into the Transformer through depth position encoding to improve detection accuracy, but it requires LiDAR-assisted supervision. MonoDETR\cite{zhang2023monodetr} predicts foreground depth maps using foreground object labels to achieve depth guidance. DETR3D\cite{wang2022detr3d} introduces a Transformer architecture to handle 3D object detection tasks in multi-view images. It utilizes a 3D-to-2D query mechanism, establishing associations between multi-view features and queries through a cross-attention mechanism, thereby achieving accurate localization of 3D objects.

\subsubsection{Summary and Analysis}
Monocular 3D object detection technology has made significant progress in recent years, mainly driven by direct regression methods and data augmentation methods. Direct regression methods efficiently predict 3D object parameters through end-to-end neural networks, but there is still room for improvement in complex scenes. Data augmentation methods, by generating depth maps combined with monocular images, have significantly improved detection accuracy, albeit at the cost of increased computational complexity. Transformer-based methods leverage global attention mechanisms and multi-view feature fusion, achieving higher detection accuracy and robustness, but they face challenges in terms of high computational complexity and inference speed. Future research needs to focus on enhancing detection performance while addressing issues of real-time processing and computational resource requirements.

%%==================================================================================================================
\subsection{Based on LiDAR for 3D Object Detection}
\begin{figure}[t]
\centering
\includegraphics[width=0.8\linewidth]{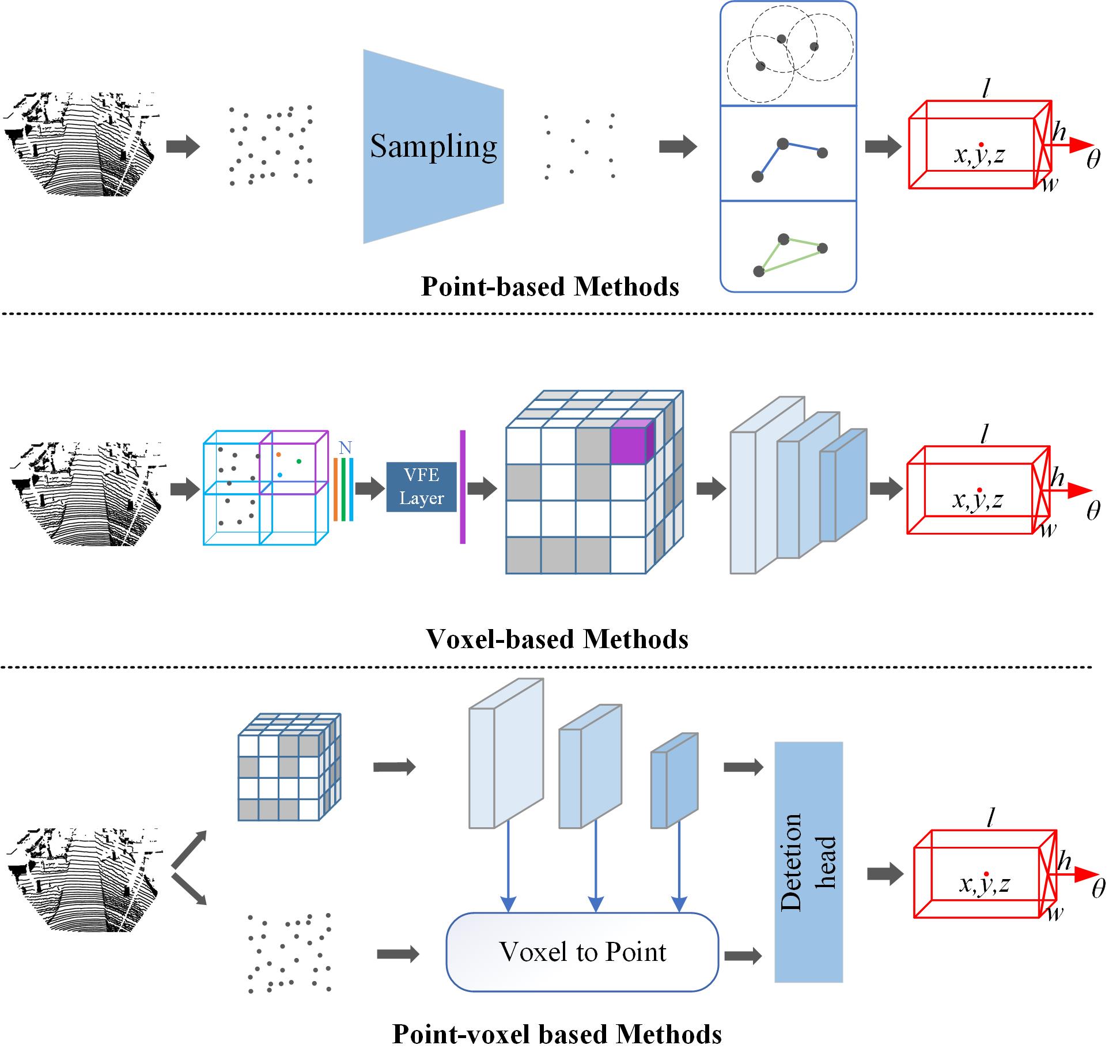} % 调整图片宽度为页面宽度的一半
\caption{An illustration of Point Cloud based Methods.}
\label{fig4}
\end{figure}

LiDAR sensors work by emitting laser pulses and measuring the time difference between these pulses and their reflection from objects in the environment. This time difference is used to calculate the distance of the laser beam to the object, thereby generating high-precision 3D point cloud data. This data provides a detailed description of the physical environment surrounding the vehicle, offering critical information for advanced driver assistance systems. Compared to cameras or radar, LiDAR can operate consistently under various lighting conditions since it does not rely on external light sources. As illustrated in Figure \ref{fig4}, this technology can directly measure the distance to surrounding objects without requiring complex algorithms to infer depth, resulting in higher accuracy for detection tasks compared to camera-only systems. However, LiDAR systems also face some limitations in practical applications, with point cloud sparsity being a notable issue\cite{bilik2022comparative}. Point cloud sparsity arises mainly from two factors: first, as distance increases, the energy of the laser pulses attenuates, weakening the intensity of the reflected signal, which reduces the amount of point cloud data for distant targets and decreases accuracy. Second, due to the inherent divergence of laser beams, the angular resolution of LiDAR causes the spatial distribution of scanned points to increase with distance, making the point cloud data more sparse for distant objects. Additionally, LiDAR's insensitivity to color and texture limits its effectiveness in applications that require detailed environmental understanding and object recognition.

\subsubsection{Point-based Methods}
Point-based methods directly process raw point cloud data without other forms of spatial transformations, which is the most intuitive approach. However, due to the direct handling of large numbers of points, they are often accompanied by down-sampling (e.g., FPS algorithm) and other processes to reduce the number of low-quality points\cite{orts2013point,zou2020point,nezhadarya2020adaptive}.Point-GNN\cite{shi2020point} is an innovative graph neural network approach for precise object detection from sparse 3D point cloud data. This method constructs a graph structure of the point cloud and performs multi-layer message passing and feature aggregation on it, allowing each point's feature representation to capture both local and global geometric information. Ultimately, each point predicts a 3D bounding box through a regression network. This approach has significant advantages in handling unstructured point clouds, effectively capturing local and global structural information, but it can become computationally and memory intensive when dealing with very dense point cloud data.Point2Seq\cite{xue2022point2seq} is a simple and effective framework for 3D object detection from point clouds. This framework treats each 3D object as a sequence of words and reformulates the 3D object detection task as decoding words from the 3D scene in an autoregressive manner. Point2Seq significantly outperforms previous anchor-based and center-based 3D object detection frameworks without any additional features.\cite{chen2024cross} introduces an innovative method that enhances the accuracy and efficiency of 3D object detection by shifting point cloud features across clusters. This approach excels in handling complex scenarios such as occlusions and sparse point clouds, significantly enhancing the model's discriminative power while reducing computational costs, demonstrating its potential application in autonomous driving.

\subsubsection{Voxel-based Methods}
Voxel-based methods simplify the data processing workflow by converting continuous point cloud data into discrete voxel grids\cite{liu2019point}. This approach is significantly influenced by 2D image processing techniques, such as regularizing irregular point cloud data into regular voxel grids and compressing extracted features into two-dimensional feature matrices. Specifically, this method involves dividing the space into regular three-dimensional grids and aggregating specific features of the point cloud within each voxel, such as the number of points, the center position, or other statistical information. In this approach, the use of 3D convolutional neural networks (CNNs) is particularly critical, as this network structure is especially suited for learning spatial hierarchies from voxel representations. For example, VoxelNet\cite{zhou2018voxelnet} is a pioneering voxel-based network that uses 3D CNNs to identify and classify objects in three-dimensional data. The core innovation of VoxelNet lies in its integration of feature encoding and object detection processes, demonstrating superior performance in multiple benchmarks. However, a major drawback of voxelization methods is that the computational cost increases significantly with voxel resolution, and detailed information may be lost during the voxelization process. To address this issue, SECOND\cite{yan2018second} has further developed networks based on sparse convolution, proposing 3D Sparse Convolution (3D SpConv) and 3D Submanifold Sparse Convolution (3D Sub-SpConv) network models. These effectively address the sparsity problem of point clouds and accelerate convolution operations. Additionally, these operations are performed only on non-empty voxels, significantly reducing the necessary computational resources and memory usage, enabling the model to handle higher resolution data. VoxelNext\cite{chen2023voxelnext} further advances 3D CNNs by utilizing the sparse nature of point clouds, extracting features through 3D CNNs, and predicting 3D boxes directly on sparse features without converting to dense feature maps, relying purely on 3D CNNs without the need for sparse-to-dense conversion or NMS post-processing. SAFDNet\cite{zhang2024safdnet} has designed an adaptive feature diffusion strategy to address the common issue of central feature loss in sparse feature detectors, effectively solving the problem of high computational cost in long-range detection for existing high-performance 3D object detectors.

\subsubsection{Point-voxel based Methods}
Point-voxel methods combine the direct processing of point cloud data with the efficiency of voxelization. These methods are notable for their ability to integrate the detailed level of point clouds with the computational efficiency of voxel representation. PV-RCNN\cite{shi2020pv} enhances detection performance by integrating point cloud features with voxel features. It employs voxelization to obtain coarse spatial features and then utilizes the Voxel Set Abstraction (VSA) module to fuse multi-scale point cloud features with voxel features, enabling the model to better understand the 3D structure of the scene. VoxSeT\cite{he2022voxel} manages and processes voxelized point clusters through an innovative Voxel-based Set Attention (VSA) module, improving computational efficiency. This method reduces the computational demand of self-attention within each voxel using a cross-attention mechanism and can process voxels of different sizes in parallel with linear complexity. Through this point-voxel fusion strategy, VoxSeT significantly optimizes the efficiency of handling large-scale point cloud data while maintaining high performance. BADet\cite{qian2022badet} improves detection accuracy and efficiency by constructing a local neighborhood graph to associate various proposal regions. BADet also introduces a lightweight regional feature aggregation module that fully leverages voxel-level, pixel-level, and point-level features, providing richer information for each region of interest.

\subsubsection{Summary and Analysis}
LiDAR-based 3D object detection technology plays a crucial role in autonomous driving and advanced driver assistance systems. LiDAR, with its stable performance under various lighting conditions and high-precision distance measurement capability, significantly outperforms traditional camera and radar systems. However, challenges such as point cloud sparsity and high computational costs remain prevalent.Recent research has introduced sparse convolution techniques and point-voxel fusion methods, which, through multi-scale feature fusion and cross-attention mechanisms, have significantly improved detection performance and computational efficiency. For example, methods like PV-RCNN and VoxSeT excel in handling large-scale point cloud data by optimizing the use of computational resources, thus addressing performance bottlenecks in high-density scenarios.

Future development directions should focus on the following points: first, improving real-time performance and computational efficiency through algorithm optimization and hardware acceleration; second, multi-sensor fusion, integrating LiDAR with camera and radar data, to further enhance detection accuracy and robustness; and third, focusing on solving the reliability issues in long-range detection and complex environments. Continued innovation and cross-disciplinary collaboration will be crucial for the advancement of LiDAR-based 3D object detection technology.
%%===================================================================================================================
\subsection{Fusion-Based 3D Object Detection}
In the field of 3D object detection, methods based on multi-sensor data fusion have become a key technology for improving detection accuracy and robustness. This approach integrates data from various sensors such as LiDAR and cameras, leveraging the complementary advantages of each sensor to achieve precise environmental perception. Although this high-precision fusion technology significantly enhances perceptual capabilities, it also poses high demands on data processing and storage resources, which is a considerable challenge for real-time applications such as autonomous vehicles. The difference in data types increases the complexity of data synchronization and processing, potentially leading to increased system latency and computational costs. Since current 3D object detection methods show that LiDAR-based approaches perform much better than image-based methods, we consider the LiDAR channel as the primary channel. Depending on the type of auxiliary information from the image channel, fusion methods can be divided into three types: data-level fusion, feature-level fusion, and result-level fusion. Each fusion method has its unique advantages and applicable scenarios, and choosing the right fusion strategy is crucial for optimizing system performance.
\begin{figure}[t]
\centering
\includegraphics[width=1\linewidth]{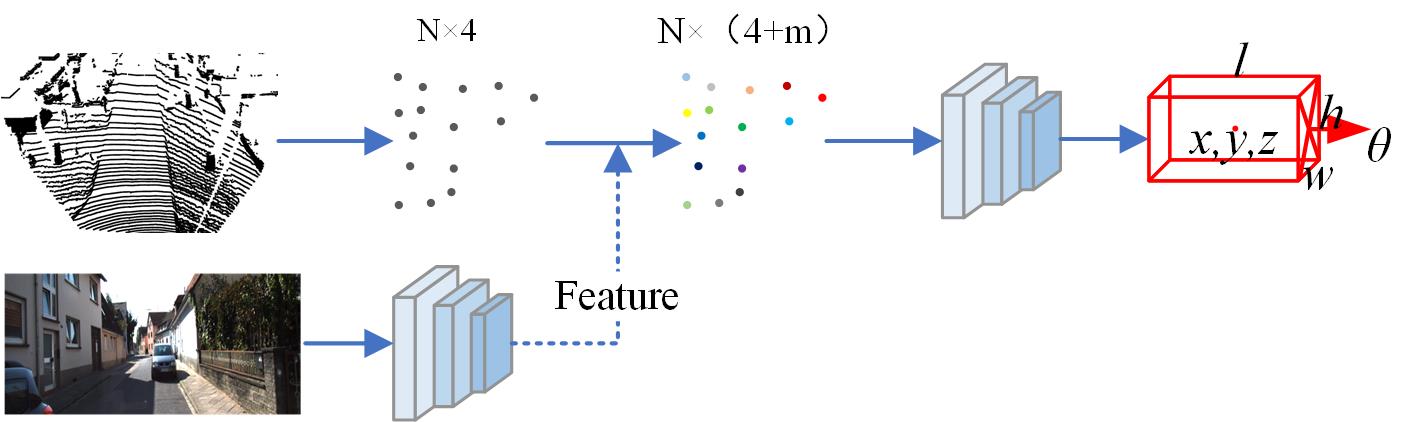} % 调整图片宽度为页面宽度的比例
\caption{An illustration of Data-Level fusion.}
\label{fig5}
\end{figure}
\subsubsection{Data-Level Fusion}
In 3D object detection, data-level fusion aims to maximize the retention of original environmental details, providing a comprehensive view of objects' positions, shapes, and textures. As shown in Figure \ref{fig5}, for example, PointPainting\cite{vora2020pointpainting} is a representative method that first applies a semantic segmentation algorithm to images captured by a camera to identify the classes of each pixel (such as pedestrians, vehicles, etc.). These semantic labels are then "painted" onto the corresponding LiDAR point cloud, so each point in the point cloud contains not only spatial coordinates and reflectivity information but also rich semantic information. MVX-Net\cite{sindagi2019mvx} proposes two different early fusion methods: PointFusion and VoxelFusion. PointFusion projects 3D points onto the image plane and concatenates 2D image features extracted by a pre-trained network to each 3D point's features, enabling the learning of useful information from both modalities simultaneously. VoxelFusion performs fusion at the voxel feature encoding layer, integrating image features into the corresponding voxel and concatenating them with the voxel's point cloud features. This method can be extended to voxels without points, improving fusion flexibility. F-PointNet\cite{qi2018frustum} utilizes a 2D object detector to generate 2D bounding boxes around objects of interest, uses a calibration matrix to project objects within the 2D boxes into 3D frustum spaces, and then processes the point cloud within the frustum for 3D detection.

\begin{figure}[t]
\centering
\includegraphics[width=1\linewidth]{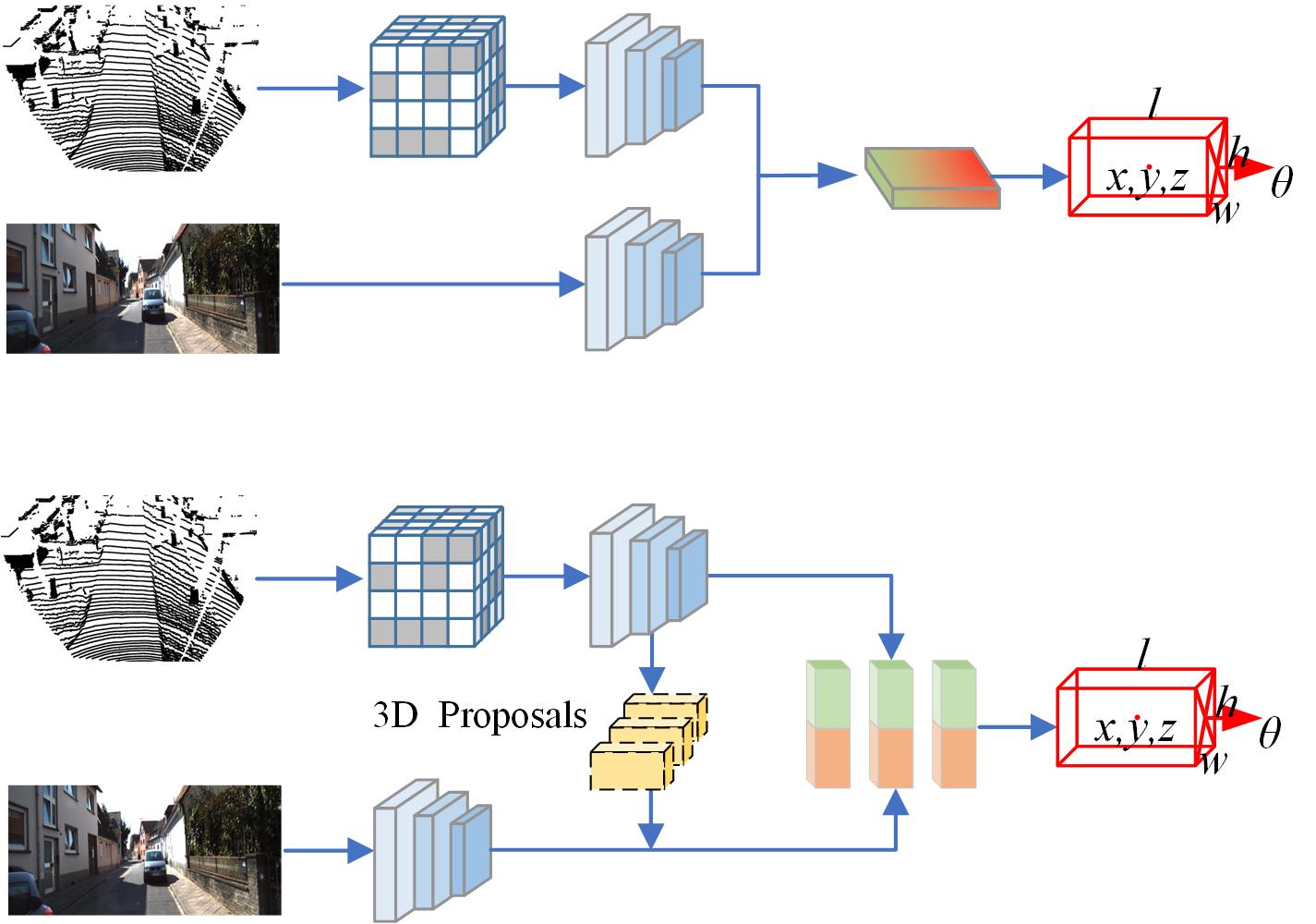} % 调整图片宽度为页面宽度的比例
\caption{An illustration of Feature-Level fusion.}
\label{fig6}
\end{figure}

\subsubsection{Feature-Level Fusion}
Feature-level fusion in 3D object detection integrates feature representations from different sensors to achieve data fusion, ensuring sufficient environmental perception details while maintaining data processing efficiency. SFD\cite{wu2022sparse} uses depth completion techniques to generate pseudo point clouds and then fuses features extracted from both the original and pseudo point clouds to achieve better detection results. 3D-CVF\cite{yoo20203d} employs attention maps to adaptively fuse semantics from point clouds and associated images. LoGoNet\cite{li2023logonet} is an innovative LiDAR-camera fusion network that effectively performs data fusion at both local and global levels. This method combines strategies of Global Fusion (GoF) and Local Fusion (LoF) to overcome the shortcomings of traditional global fusion methods in handling fine-grained region-level information, thereby improving detection accuracy and efficiency. The core idea of BEVFusion\cite{liang2022bevfusion} is to fuse data from different sensors into a unified representation space in the Bird's Eye View (BEV). Camera and LiDAR data are processed independently, and their features are then fused in the BEV space. This design allows the system to retain functionality even when data from one sensor is unavailable. IS-FUSION\cite{yin2024isfusion} significantly enhances 3D object detection performance by combining instance-level and scene-level multi-modal information. Unlike existing methods that focus only on BEV scene-level fusion, IS-FUSION explicitly incorporates instance-level multi-modal information.As shown in Figure \ref{fig6}, BEVFusion is a representative of the first category of methods, while SFD and LoGoNet are examples of the second category.
\begin{figure}[t]
\centering
\includegraphics[width=1\linewidth]{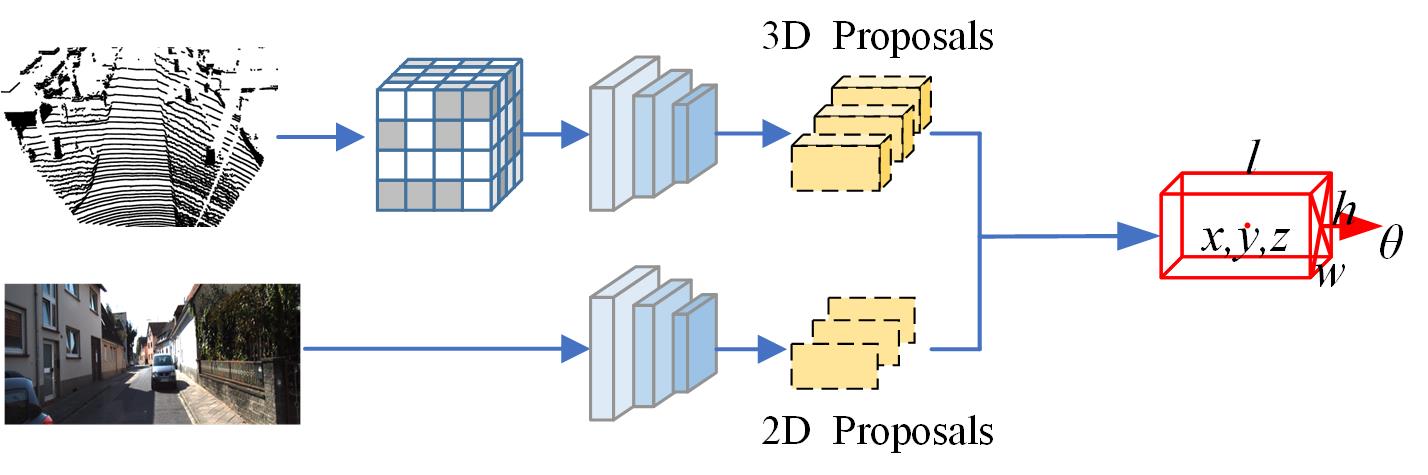} % 调整图片宽度为页面宽度的比例
\caption{An illustration of Result-Level fusion.}
\label{fig7}
\end{figure}

\subsubsection{Result-Level Fusion}
Result-level fusion is an efficient strategy that merges the output results of various detection algorithms after they complete their respective tasks. The main advantage of this approach lies in its high flexibility and ease of implementation, allowing each detection system to operate independently, with final results integrated through specific logical or statistical methods. In multi-sensor systems, even if the output of one sensor is compromised, this method can still achieve reliable final judgments using results from other sensors, providing additional robustness. As shown in Figure \ref{fig7},CLOCs\cite{pang2020clocs} (Camera-LiDAR Object Candidates) is a typical example of result-level fusion, optimized specifically for combining detection results from cameras (2D) and LiDAR (3D). Its core idea is to cross-validate candidate objects from 2D and 3D detectors before performing Non-Maximum Suppression (NMS), leveraging their geometric and semantic consistency to generate more accurate final detection results. This strategy not only enhances detection accuracy but also significantly improves performance in complex environments.

\subsubsection{Summary and Analysis}
In 3D object detection, multi-sensor data fusion methods have significantly improved detection accuracy and robustness. Data-level fusion maximizes the retention of original environmental details, providing a comprehensive view for detection; feature-level fusion integrates feature representations from different sensors, balancing data processing efficiency with perceptual accuracy; and result-level fusion merges the output results of various detection algorithms, offering highly flexible and straightforward implementation strategies.

While data-level fusion methods like PointPainting significantly improve detection accuracy, they demand high computational resources and data processing capabilities. Feature-level fusion methods such as LoGoNet achieve more efficient 3D object detection by integrating multi-modal features, overcoming traditional methods' limitations in handling fine-grained information. Result-level fusion methods like CLOCs enhance system robustness and accuracy by merging detection results from different sensors, excelling in complex environments.

Future development should focus on optimizing fusion algorithms to improve real-time performance and reduce computational resource consumption, strengthening the synchronization and collaborative processing of multi-sensor data to enhance overall system performance, and developing smarter fusion strategies to address complex application scenarios. Continued innovation and optimization in multi-sensor data fusion will play a more significant role in 3D object detection, advancing autonomous driving and intelligent systems.
\section{Future Development Directions}

\subsection{Temporal Perception Methods}

Humans naturally possess the ability to predict the future, whether in daily life or while driving, because the world is dynamic, not static. However, current object detection methods typically rely on single-frame point cloud or image data. Many recent studies have shown that utilizing temporal sequence data can significantly improve detection accuracy and robustness. By capturing the dynamic changes of objects and the continuity over time, temporal 3D object detection methods can more comprehensively understand the surrounding environment, avoiding occlusion and detection failures. This section introduces the technical principles and major advancements of 3D object detection methods based on LiDAR sequences and image sequences, respectively.

\subsubsection{Real-time Temporal Perception}
\begin{figure}[t]
\centering
\includegraphics[width=0.8\linewidth]{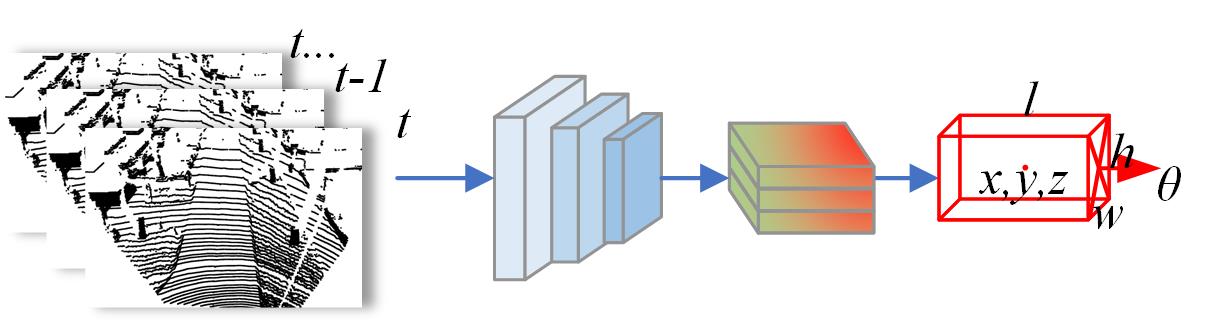} 
\caption{An illustration of Real-time Temporal Perception.}
\label{fig8}
\end{figure}

\textbf{LiDAR Sequence 3D Object Detection:}As illustrated in Figure \ref{fig8}, LiDAR sequence-based 3D object detection leverages multi-frame point cloud data to capture richer spatial information and temporal dynamics, effectively improving detection accuracy and robustness. Multi-frame point cloud fusion methods, such as Spatiotemporal Networks and sliding window methods, accumulate and process point cloud data from multiple consecutive time frames, enhancing spatial resolution and detection accuracy\cite{luo2018fast}. Additionally, temporal modeling tools like RNNs (Recurrent Neural Networks), LSTMs (Long Short-Term Memory networks)\cite{huang2020lstm}, GRUs (Gated Recurrent Units)\cite{yin2020lidar}, and Transformers excel in capturing temporal dynamic changes.For example, STINet\cite{zhang2020stinet} improves accuracy by jointly modeling spatial and temporal information, generating temporal proposals to extract geometric and dynamic features of targets, and modeling interactions between pedestrians through interactive layers. These methods have demonstrated significant effectiveness in practical applications. Yin et al\cite{yin2020lidar}. proposed a method using graph networks and temporal attention mechanisms , while Yuan et al\cite{yuan2021temporal}. introduced the Temporal-Channel Transformer, utilizing a temporal-channel encoder to capture complex interactions between different frames and channels, and a spatial decoder to precisely enhance feature representations of target frames, significantly improving detection accuracy and speed.PTT (Point-Trajectory Transformer)\cite{huang2024ptt} proposed an efficient temporal 3D object detection framework that does not require multi-frame point clouds. By using only the current frame's point cloud and multi-frame proposal trajectories, it minimizes memory requirements. This method introduces long short-term memory and future encoding modules, enhancing feature extraction capabilities.

\textbf{Image Sequence 3D Object Detection:}In the field of autonomous driving and intelligent transportation systems, 3D object detection technology based on image sequences is rapidly developing, using temporal information from consecutive frames to improve detection accuracy. Optical flow methods\cite{menze2015object} capture the dynamics of objects by analyzing motion vectors between pixels, providing key information for scene understanding, as shown in \cite{hu2019joint}. They effectively link detection results between frames by combining object detection and tracking, enhancing the coherence of target tracking. At the same time, temporal convolutional networks (TCNs)\cite{bai2018empirical} enhance the stability of detection by deeply mining the temporal features of image sequences through one-dimensional convolution. Recurrent neural networks, such as LSTM and GRU, further capture dynamic changes by analyzing frame by frame, enhancing detection accuracy. In addition, the algorithm proposed by Zeng et al\cite{zeng2022lift}. significantly enhances the accuracy and robustness of detection by fusing time series data with a deep learning model. Recently, the transformer architecture has also been introduced into this field. Models such as BEVFormer\cite{li2022bevformer} and PETRv2\cite{liu2023petrv2} use spatiotemporal attention mechanisms to effectively transform from multiple camera perspectives to a unified bird's-eye view (BEV), providing a new solution for all-round environmental perception. The development of these technologies has not only advanced 3D object detection technology but also provided strong support for the safety and reliability of autonomous driving. StreamPETR\cite{wang2023exploring} uses an object-oriented temporal mechanism, propagating long-term historical information by propagating object queries frame by frame in multi-perspective video frames. This method effectively captures the motion and changes of target objects over time.

\subsection{Predictive Temporal Perception}
\begin{figure}[t]
\centering
\includegraphics[width=0.8\linewidth]{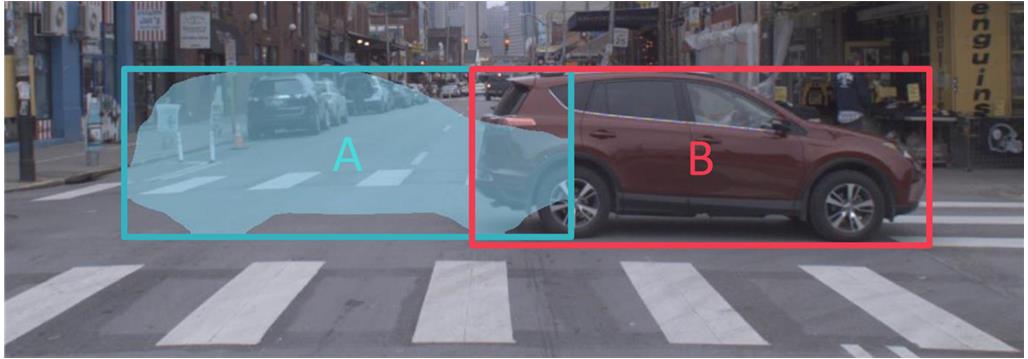} 
\caption{An illustration of Predictive Temporal Perception.}
\label{fig9}
\end{figure}

Predictive Temporal Perception(Streaming perception) tasks refer to the consideration of both latency and accuracy in a continuous manner. As shown in Figure \ref{fig9}, the methods that utilize historical and current frames to predict future moments in the environment are collectively referred to as predictive temporal perception. The work Towards Streaming Perception\cite{li2020towards} first introduced the concept of streaming AP (sAP) to evaluate accuracy while considering latency, highlighting that non-real-time detectors may miss certain frames. In reality, whether it’s a real-time detector or a non-real-time detector, there will always be at least one unit of time delay, and the impact of this delay fluctuates with the speed of the detector and the changing dynamics of the surrounding environment. To overcome this challenge, some researchers have developed predictive perception models based on real-time detection. These models explicitly predict the future using learning-based approaches rather than merely seeking a better trade-off between accuracy and latency, thereby further improving performance.

This task was first explored in the context of semantic segmentation. Recent studies have transformed this task into the prediction of intermediate segmentation features using techniques such as deformable convolutions\cite{chiu2020segmenting}, teacher-student learning\cite{vsaric2019single}, flow-based prediction\cite{saric2020warp}, and LSTM methods\cite{lin2021predictive}. Subsequently, some methods have applied this approach to object detection tasks, such as StreamYOLO\cite{yang2022streamyolo} and LongShortNet\cite{li2023longshortnet}, which predict the next time step’s object detection based on real-time detectors using learning-based methods. In the field of 3D object detection, some researchers have begun to explore this approach. For example, \cite{yu2023vehicle} via Feature Flow Prediction and \cite{yu2024flow} have applied this method to reduce latency in vehicle-infrastructure cooperative 3D object detection. By predicting future features of roadside equipment, they aim to mitigate the effects of communication delays.

\subsection{Summary and Analysis}

Temporal 3D object perception methods leverage consecutive frames of LiDAR and camera data, aligning more closely with how humans perceive dynamic environments. LiDAR temporal perception uses consecutive frames of point cloud data to integrate spatial information over time, achieving precise identification of object positions and motion states. On the other hand, camera temporal perception captures motion trajectories and dynamic changes using consecutive frames of image data, enhancing 3D object detection performance. However, processing multi-frame inputs involves high computational complexity and consumes significant computational power. To address this challenge, current research proposes using previous object detection results or trajectory features to replace multi-frame point cloud data, thereby reducing computational load. These strategies not only maintain high detection accuracy but also significantly improve computational efficiency, making temporal 3D object perception methods more practical for real-time applications such as autonomous driving.

\subsection{3D Occupancy Grid}

Quantifying 3D scenes into structured units with semantic labels, known as 3D occupancy, brings significant advantages to autonomous driving. Traditional 3D object detection techniques can determine the position of objects such as vehicles but often simplify the object's contour to a 3D box. While this method describes the general spatial location of objects, it fails to capture more detailed structures. In contrast, 3D occupancy not only clarifies the spatial position of objects but also retains detailed structural information. This technique integrates the advantages of existing description methods, making it highly suitable for the autonomous driving field and has already been adopted in the industry by companies like Tesla and Mobileye. Consequently, the significant advantages of 3D occupancy have prompted researchers to explore its potential in enhancing traditional perception tasks and optimizing downstream planning processes.

\begin{figure}[t]
\centering
\includegraphics[width=0.8\linewidth]{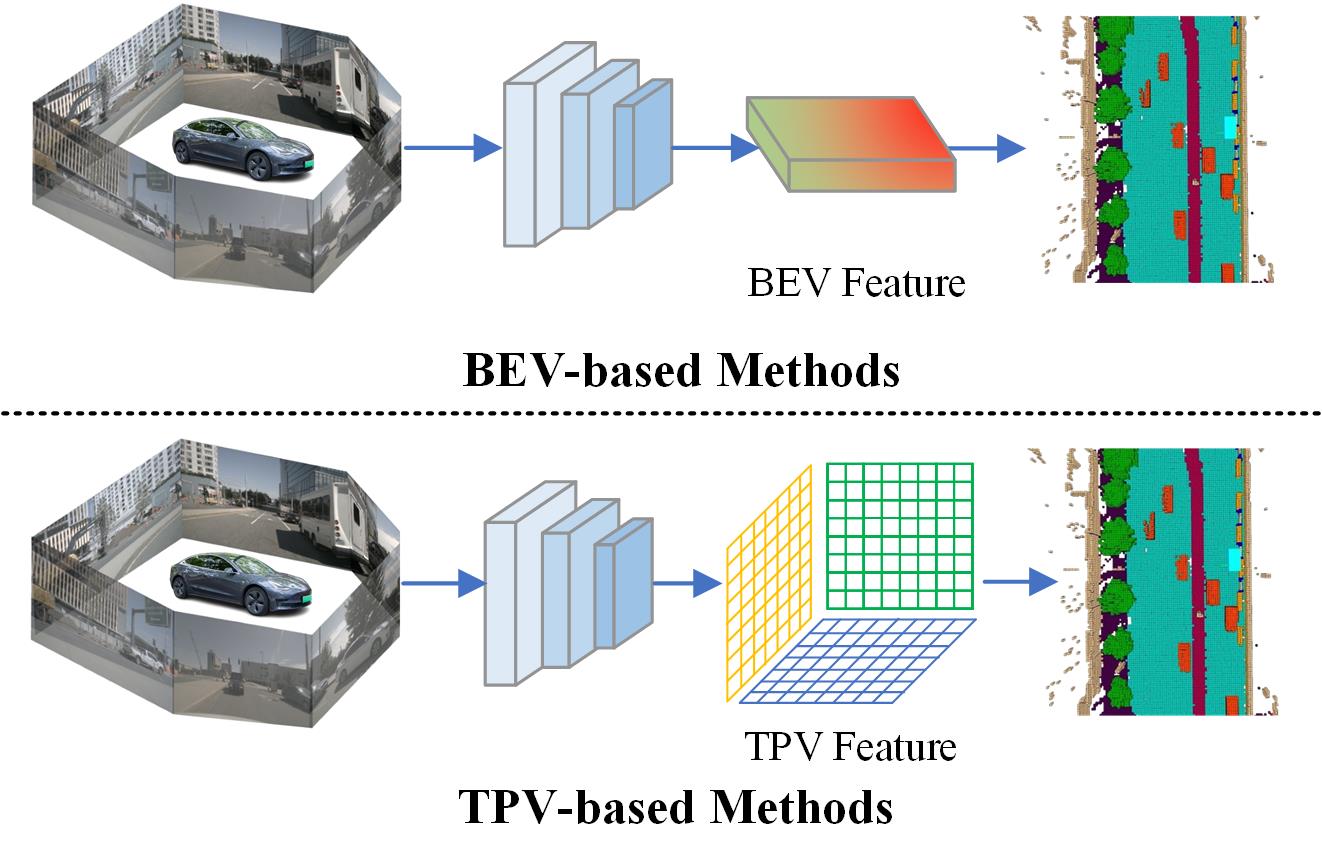} 
\caption{An illustration of 3D Occupancy.}
\label{fig10}
\end{figure}

\subsubsection{LiDAR-Based 3D Occupancy}
\begin{table}[h!]
\centering
\caption{Comparison of 3D Occupancy Prediction with Other Methods.}
\label{table2}
\resizebox{\textwidth}{!}{
\begin{tabular}{|l|l|l|c|l|l|l|}
\hline
\textbf{Technical Type} & \textbf{Accuracy} & \textbf{Output Format} & \textbf{Background \& Mapping} & \textbf{Real-time Capability} & \textbf{Environment Understanding} \\ \hline
3D Occupancy Perception & High & 3D & $\checkmark$ & Medium & High \\ \hline
3D Bounding Box Detection & Medium & 3D & $\times$ & High & Low \\ \hline
BEV Segmentation & Low & 2D & $\checkmark$ & High & Medium \\ \hline
Point Cloud Segmentation & High & 3D & $\checkmark$ & Medium & High \\ \hline
\end{tabular}}
\end{table}

In the field of autonomous driving, LiDAR-based 3D occupancy technology provides a more detailed and comprehensive understanding of the environment. By performing semantic occupancy prediction on point cloud data, not only the position of objects can be determined, but their fine structures can also be captured. To achieve this goal, researchers have proposed various innovative methods. The PointOcc\cite{zuo2023pointocc} model introduces a novel representation for point cloud data through the Cylindrical Tri-Perspective View (Cylindrical TPV). This perspective enhances comprehensive description of the point cloud and, by integrating with the processing power of 2D backbone networks, effectively models 3D information, addressing key issues in 3D semantic occupancy prediction.Furthermore, the DIFs\cite{rist2021semantic} propose an innovative approach based on Local Deep Implicit Functions for scene segmentation. This method represents LiDAR data with continuous functions, bypassing the spatial discretization limitations of traditional voxelization methods, achieving local encoding of raw point clouds, and precise reconstruction of the global scene.OCF\cite{liu2023lidar} introduces a novel LiDAR perception task—Occupancy Completion and Forecasting (OCF). By unifying scene completion and occupancy prediction, this approach addresses challenges such as sparse-to-dense reconstruction, partial-to-complete hallucination reasoning, and spatial-to-temporal dimension prediction. The researchers developed the OCFBench dataset and evaluated baseline models, with results showing that this new method performs exceptionally well in handling complex dynamic environments, providing a new direction for 4D perception research.

These studies have not only achieved breakthroughs in their respective technologies but also complement each other, jointly promoting the development of environmental perception technology in the field of autonomous driving. Through these methods, autonomous driving systems can understand their surroundings more finely, providing a solid foundation for safe and reliable driving decisions.As shown in Table \ref{table2}, these methods vary in their ability to understand the environment, highlighting the different levels of detail and accuracy they can achieve.

\subsubsection{Camera-Based 3D Occupancy}

Compared to point cloud data, image data carries more features aligned with human visual perception and has a relatively smaller data volume. However, effectively performing occupancy perception in images lacking depth information has become a key focus for researchers. MonoScene\cite{cao2022monoscene}, as a pioneering technology, achieves 3D semantic scene modeling and understanding using only a monocular camera, based on a single RGB image. Unlike traditional methods that rely on depth or multi-view data, MonoScene utilizes a series of 2D and 3D networks and a feature projection technique based on optical principles (FLoSP) to achieve accurate 2D-to-3D conversion. TPVFormer\cite{huang2023tri} takes a different approach by introducing a three-view (TPV) representation to address the view representation challenges in 3D semantic occupancy prediction. TPVFormer combines a bird’s-eye view with two vertical views, using a Transformer encoder to extract TPV features from 2D images and applying attention mechanisms to enable feature interaction between views, thereby providing a more comprehensive description of 3D scenes.Figure \ref{fig10} illustrates two mainstream approaches.

Moreover, researchers have explored other efficient representation methods. For example, ViewFormer\cite{li2024viewformer} proposes a multi-view 3D occupancy perception model based on a View-Guided Transformer, which improves perception accuracy and efficiency through spatiotemporal modeling. GaussianFormer\cite{huang2024gaussianformer} enhances visual 3D semantic occupancy prediction by representing the scene as a Gaussian distribution, making the modeling more precise.

Complementing these methods, OccNet\cite{tong2023scene} introduces a multi-view vision-centric framework that combines a cascaded voxel decoder and temporal self-attention mechanism to reconstruct 3D occupancy representations. The core innovation of OccNet lies in its general occupancy embedding, which makes the generation of voxel features from image features more efficient and widely applicable. PanoOcc\cite{wang2024panoocc} further proposes a unified occupancy representation method for camera-based 3D panoramic segmentation, effectively improving the system’s overall performance and generalization capability.

In the realm of self-supervised learning, SelfOcc\cite{huang2024selfocc} uses self-supervised strategies to generate high-quality occupancy predictions even in the absence of explicit annotations, further enhancing the accuracy of vision-based 3D occupancy perception. Similarly, Cam4DOcc\cite{ma2024cam4docc} introduces a benchmark for 4D occupancy prediction with cameras, incorporating the temporal dimension to predict future scene changes, significantly improving occupancy perception in dynamic environments.

\subsubsection{Multi-Modal Fusion-Based 3D Occupancy}

Multi-modal fusion has demonstrated significant advantages in 3D semantic occupancy prediction, as it can integrate features from different sensors to enhance the model's accuracy, robustness, and generalization capability. OpenOccupancy\cite{wang2023openoccupancy} provides a comprehensive benchmark and establishes baselines for camera-based, LiDAR-based, and LiDAR-camera-based approaches. Some works have incorporated even more sensors to achieve higher precision. OccFusion\cite{ming2024occfusion}, for instance, uses dynamic 3D/2D fusion modules to integrate features from different modalities, leveraging multi-level supervision mechanisms to improve the model's robustness and accuracy. Experimental results show that OccFusion performs exceptionally well in challenging environments such as rainy and nighttime conditions. LiCROcc\cite{ma2024licrocc} proposes a method that uses LiDAR and camera data to teach radar for accurate semantic occupancy prediction. By combining the rich information from LiDAR and cameras with the cost-effectiveness of radar, this method significantly enhances radar's performance in 3D occupancy prediction tasks. However, 3D occupancy is inherently resource-intensive, and the use of multi-modal fusion further increases computational demands. Therefore, despite its effectiveness, this approach has not been widely studied.

\subsubsection{Summary and Analysis}

3D occupancy technology holds immense potential in the field of autonomous driving, significantly advancing environmental perception capabilities. By quantifying and semantically annotating three-dimensional scenes, this technology not only determines the position of objects but also captures their detailed structures, providing autonomous driving systems with more precise and enriched environmental information. Methods based on LiDAR and cameras each demonstrate unique advantages, such as innovations in point cloud representation and monocular image processing. These technologies have laid a solid foundation for future research, proving that integrating multi-sensor data can substantially enhance system robustness and accuracy.

However, the fusion of multimodal data also introduces increased computational complexity. Balancing high performance with reduced computational costs is a critical challenge that future research must address. For instance, the successful application of techniques like OccFusion, which achieve excellent performance in complex environments, hinges on the use of cleverly designed fusion strategies. Future research should focus on optimizing these fusion strategies and exploring new algorithms and architectures to achieve more efficient and real-time 3D occupancy technology.

\subsection{End-to-End Autonomous Driving}

End-to-end autonomous driving systems integrate the perception, decision-making, and planning functions of a vehicle into a unified neural network, directly converting sensor data into driving commands. This approach optimizes the overall performance of autonomous driving rather than focusing solely on the efficiency of individual tasks. It is important to note that the end-to-end paradigm does not necessarily imply a black box with only planning/control outputs; it can be modular with intermediate representations and outputs that are ultimately jointly optimized. This method can effectively improve model interpretability, simplify error diagnosis, and streamline the optimization process. 

In traditional autonomous driving architectures, various tasks such as 3D object detection and path planning are typically handled separately, with each module attempting to optimize its own accuracy metrics. This can lead to trade-offs in overall performance, as lack of coordination between modules may reduce the system's overall responsiveness and accuracy. The end-to-end approach not only simplifies the system architecture but also enhances processing efficiency and response speed.

Moreover, integrating large language models (LLMs) into end-to-end autonomous driving systems can further enhance the system's environmental understanding capabilities. LLMs can process complex semantic information and interpret high-level semantics in multimodal data (such as text, images, and videos). This ability can be used to understand and interpret complex situations in driving scenarios. By combining the environmental understanding capabilities of large language models, end-to-end autonomous driving systems can achieve more precise path planning and decision-making, improving safety and reliability in various complex scenarios.

\begin{figure}[t]
\centering
\includegraphics[width=1\linewidth]{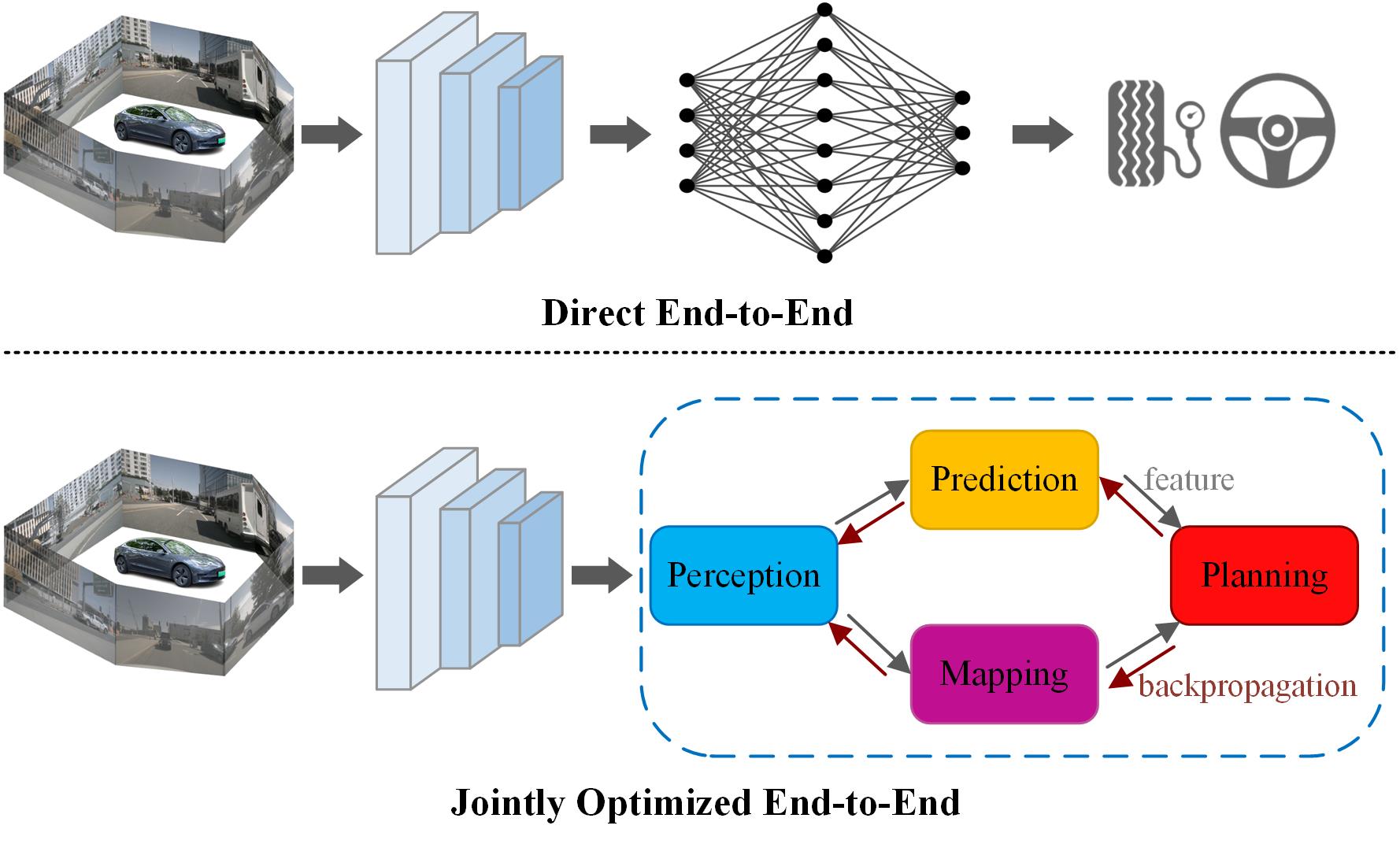} 
\caption{An illustration of End-to-End autonomous driving.}
\label{fig11}
\end{figure}

\subsubsection{Direct End-to-End}

As the name suggests, direct end-to-end autonomous driving infers vehicle operations such as steering and throttle control directly from sensor data. This approach has a simple structure but lacks interpretability. The paper\cite{bojarski2016end} proposes a method using convolutional neural networks (CNNs) to learn to drive a car from scratch, mapping raw pixels from a front-facing camera to driving commands. This end-to-end method allows the system to learn to drive on roads and highways with or without lane markings using minimal human training data. The authors found that the system could automatically learn necessary processing steps, such as detecting road features, without being explicitly trained to recognize these features, like lane lines. \cite{valiente2019controlling} uses shared image data from two self-driving cars, considering the temporal dependencies between image frames to improve the accuracy of steering angle control. The paper proposes a new deep learning architecture that integrates CNNs, long short-term memory networks (LSTMs), and fully connected layers. It not only uses the current image but also predicts the steering angle using future images shared through vehicle-to-vehicle (V2V) communication.

Imitation learning trains autonomous driving systems by mimicking expert behavior. Behavior cloning is a common method that converts expert driving behavior into a supervised learning task, training neural networks to predict vehicle control commands. Methods like DAgger (Dataset Aggregation)\cite{ross2011reduction} improve model performance by combining expert and model decisions during training\cite{chen2020learning,prakash2020exploring,zhang2017query}. Imitation learning performs well in complex urban driving scenarios but also faces issues such as bias accumulation and causal confusion\cite{li2024exploring,wen2021keyframe,park2021object}.

Reinforcement Learning (RL) learns optimal strategies through interaction with the environment, based on reward feedback. In the field of autonomous driving, RL is trained in simulated environments, gradually improving driving decision-making capabilities. Common RL algorithms include Deep Q-Networks(DQN)\cite{mnih2015human} and Policy Gradient Methods, among which Proximal Policy Optimization(PPO)\cite{schulman2017proximal} and Deep Deterministic Policy Gradient(DDPG)\cite{lillicrap2015continuous} are widely used in autonomous driving research due to their outstanding performance in policy stability and optimization efficiency\cite{kiran2021deep}.

In recent studies, the PPO algorithm has been used to optimize driving decisions under diverse road conditions. By incorporating an encoder network for image processing, this method converts sensor data into multidimensional feature vectors, enabling stable decision outputs in complex driving environments. This approach has demonstrated excellent performance in simulated environments, particularly in handling complex road scenarios\cite{wu2023proximal}. Moreover, the application of RL in heterogeneous traffic environments has been further explored. For instance, a study proposed an end-to-end autonomous driving architecture specifically designed for heterogeneous traffic, where RL is used to make driving decisions directly from sensor data. This method formulates the autonomous driving problem as a Markov Decision Process (MDP), optimizing driving performance under diverse traffic conditions and significantly enhancing the system's robustness and adaptability\cite{chakraborty2023end}.

\subsubsection{Jointly Optimized End-to-End}

In recent years, research in the field of autonomous driving has increasingly focused on generating critical safety data and advocating for modular end-to-end planning. Figure \ref{fig11} illustrates the differences between the two methods.This approach aims to address the challenges of poor interpretability and difficult optimization in traditional end-to-end models. In contrast, end-to-end models with joint optimization improve interpretability by integrating multiple tasks while significantly enhancing system efficiency and simplicity. For example,\cite{sadat2020perceive} proposes an end-to-end perception, prediction, and motion planning model. This model uses point cloud data and high-definition maps to generate safe vehicle trajectories, ensuring the interpretability of intermediate representations. Another study\cite{casas2021mp3}, introduces a map-free autonomous driving method, demonstrating strong robustness and adaptability in diverse traffic conditions. Similarly, \cite{hu2023planning} proposes a unified framework that integrates perception, prediction, and planning tasks. By using a Transformer-based decoder structure and a unified query interface, this method reduces information loss and error accumulation. In terms of multimodal data fusion, \cite{chitta2022transfuser} introduces a fusion mechanism based on a self-attention mechanism. This method captures the global 3D scene context, with particular attention to dynamic agents and traffic lights, significantly improving driving performance in complex traffic scenarios. This approach not only demonstrates the potential of multimodal fusion but also provides new directions for future research.

GenAD\cite{zheng2024genad} is a representative study that introduces generative models into end-to-end autonomous driving. By utilizing generative models, GenAD produces diverse driving decisions in complex traffic environments, allowing the system to maintain good adaptability even in unseen scenarios. This approach leverages Generative Adversarial Networks (GAN) and self-supervised learning, significantly improving the robustness and flexibility of driving decisions.

\subsubsection{LLMS for Autonomous Driving}

In recent years, with the rapid development of deep learning technologies, Large Language Models (LLMs), such as GPT-4, have made significant strides in the field of Natural Language Processing (NLP). These models not only exhibit powerful semantic understanding and generation capabilities but also show potential in processing multimodal data. As autonomous driving technology continues to advance, researchers have begun integrating LLMs into autonomous driving systems to enhance their perception, decision-making, and planning capabilities, offering new approaches to addressing complex driving scenarios\cite{cui2024survey,ding2023hilm,keysan2023can}. The introduction of LLMs enables autonomous driving systems to better understand the environment through multimodal fusion. For example, LLMs are being used to combine data from cameras and LiDAR sensors to generate richer and semantically deep environmental representations. Specifically, studies have shown that LLMs can interpret traffic signs, pedestrian behaviors, and other dynamic road elements, and integrate this information with visual inputs, thereby enhancing the environmental perception capabilities of autonomous driving systems. This multimodal fusion not only improves the system’s ability to understand complex scenes but also lays the foundation for more intelligent driving decisions in the future.

Recent research, such as \cite{fu2024drive} explores the potential of using LLMs to achieve human-like driving in autonomous vehicles. Traditional optimization-based or modular autonomous driving systems face performance limitations when dealing with complex and edge-case scenarios. In contrast, LLMs overcome these challenges through common-sense reasoning, scene interpretation, and memory functions. The study demonstrates that LLMs can make reasonable decisions in closed-loop driving environments and exhibit excellent performance in complex driving scenarios. This human-like driving approach not only enhances the system's ability to handle long-tail scenarios but also provides valuable insights for the development of more intelligent autonomous driving systems in the future.The study \cite{chen2024driving} proposes a method that integrates LLMs with object-level vector modalities to enhance the explainability of autonomous driving systems. The research shows that by combining object-level vector representations with the semantic understanding capabilities of LLMs, the system can better understand the relationships between objects in driving scenarios and generate more interpretable decision-making paths.Additionally, new frameworks more suited to autonomous driving tasks are being proposed. For instance, \cite{zhou2024embodied} introduces a new framework designed to improve the understanding and navigation capabilities of autonomous driving agents in driving scenarios. Traditional Vision-Language Models (VLMs) have limitations in the two-dimensional domain, lacking spatial awareness and long-term reasoning capabilities. However, ELM addresses these challenges by introducing spatial awareness pre-training and temporal-aware token selection mechanisms, enabling better understanding and response to complex driving environments.

\subsubsection{Summary and Analysis}

While end-to-end autonomous driving technology offers a promising direction for simplifying autonomous driving systems, it still faces challenges in terms of safety, reliability, and interpretability. Due to the complexity and black-box nature of these models, end-to-end systems struggle to provide clear explanations and diagnoses when handling unforeseen complex scenarios, making it more difficult to ensure driving safety. Additionally, training these models often requires large amounts of labeled data and computational resources, making them more expensive and time-consuming compared to traditional modular approaches.

Future research may focus on several key areas: first, increasing the transparency of models so that the system can provide reasonable explanations and responses when faced with unknown situations; second, enhancing the model's ability to handle long-tail events and complex driving scenarios; and third, reducing reliance on large-scale labeled data by employing methods such as semi-supervised learning, self-supervised learning, or transfer learning to optimize the training process. Meanwhile, the introduction of LLMs brings new ideas to autonomous driving technology. LLMs can integrate multimodal data from sources such as cameras and LiDAR, generating semantically rich environmental representations that improve the system's perception and decision-making capabilities.

\subsection{Collaborative Perception in Autonomous Driving}

\begin{figure}[t]
\centering
\includegraphics[width=1\linewidth]{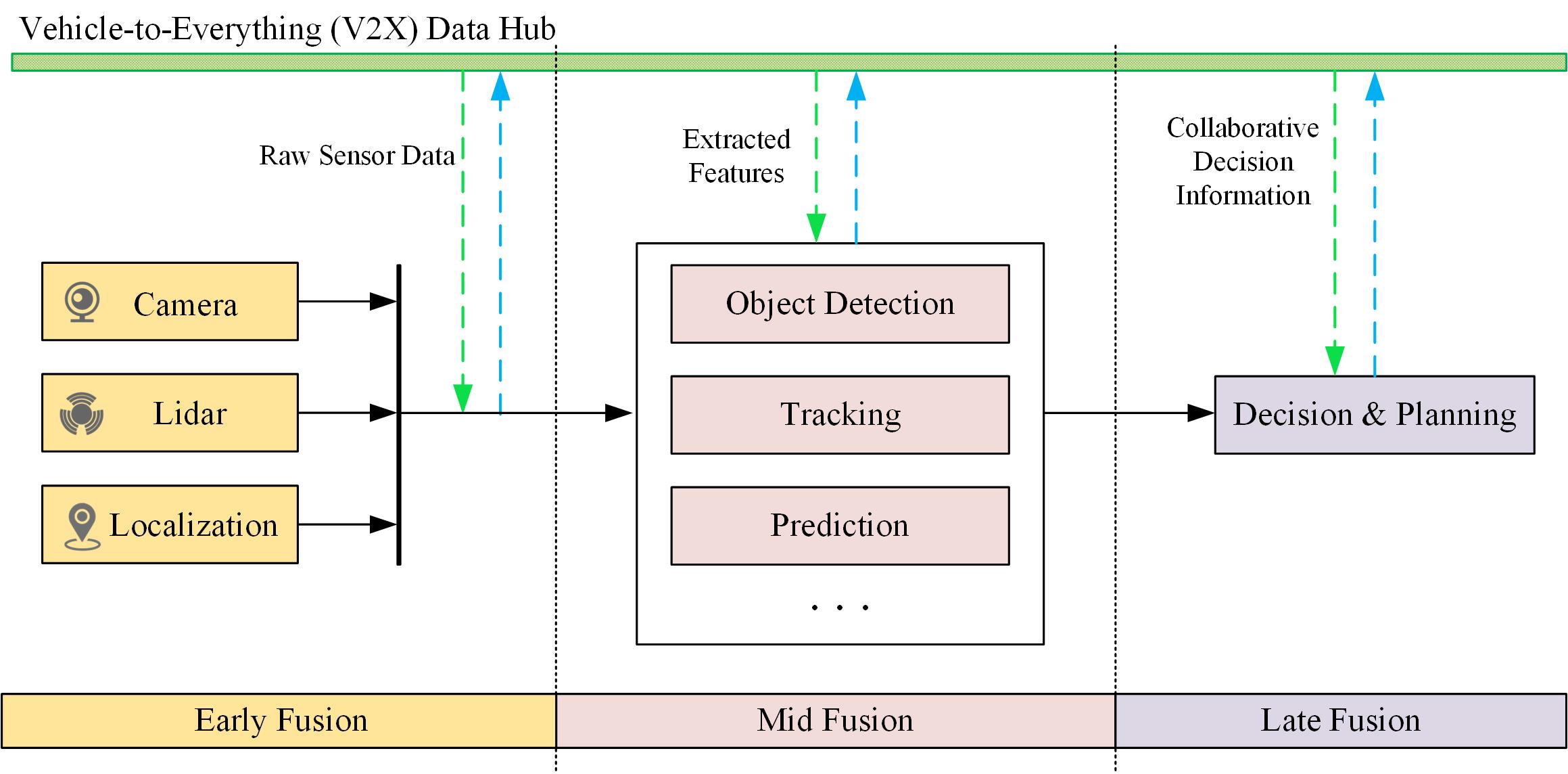} 
\caption{An illustration of collaborative perception.}
\label{fig12}
\end{figure}

In the field of collaborative perception, using Vehicle-to-Vehicle (V2V) and Vehicle-to-Infrastructure (V2I) communication to share and integrate perception information is crucial for achieving broader and more accurate environmental perception. Collectively referred to as Vehicle-to-Everything (V2X), these technologies significantly enhance the safety and efficiency of autonomous driving systems by providing comprehensive interaction between vehicles and their surroundings.As autonomous driving technology advances, the perception systems of individual vehicles face numerous challenges in complex traffic environments, especially under conditions of obstructed views or adverse weather. Collaborative perception technology integrates perception data from multiple vehicles and infrastructure, greatly expanding the perception range of a single vehicle, improving the accuracy of perception, and enhancing the overall reliability of the system. Particularly with the adoption of 5G communication technology, characterized by low latency and high data transmission speeds, the practical application of collaborative perception technology is further accelerated, making real-time data exchange and processing possible.

Similar to multi-modal sensor fusion, collaborative perception technology involves the integration of multiple data sources. However, most research primarily focuses on the fusion of the same type of sensors, such as point cloud with point cloud and image with image. Depending on the stage of fusion, it is generally divided into early fusion, mid fusion, and late fusion. At different stages of fusion, the integration of different devices, such as V2V and V2I, is also involved.Figure \ref{fig12} illustrates the overall framework of cooperative perception.

\subsubsection{V2V Communication and Collaborative Perception}

Based on the type of perception data, vehicle-to-vehicle (V2V) cooperative fusion can be mainly divided into camera-based V2V fusion and point cloud-based V2V fusion. In camera-based V2V fusion, the images captured by vehicle cameras provide rich visual information. However, due to inherent limitations of cameras, depth prediction bias becomes a major challenge. To overcome this issue, CoCa3D\cite{hu2023collaboration} leverages agent cooperation to significantly enhance 3D detection capabilities using only camera-based collaboration. The CoCa3D framework addresses this limitation by allowing multiple agents equipped only with cameras to share visual information. By sharing depth information at the same points, errors are reduced, improving the handling of depth ambiguities and extending detection capabilities to occluded and long-distance areas that are typically challenging for a single camera. Experimental results even surpass those of LiDAR-based methods.

On the other hand, point cloud-based V2V fusion primarily uses three-dimensional point cloud data generated by LiDAR and other devices to provide high-precision spatial information. However, the large volume of point cloud data, along with transmission and processing delays and noise, poses major obstacles to achieving real-time cooperative perception. To address this issue, CodeFilling\cite{izadi2022codefill} optimizes the representation and selection of cooperative messages, significantly improving the trade-off between perception and communication. Specifically, it adopts a codebook-based message representation method, allowing the transmission of integer codes instead of high-dimensional feature maps. Additionally, the proposed information-filling-driven message selection method optimizes local messages to collectively fulfill each agent's information needs, preventing information overflow between multiple agents.MRCNet\cite{hong2024multi} reduces communication pressure by filtering features based on information content and addresses real-world environmental noise issues that are often overlooked in current research, such as pose noise, motion blur, and perception noise. The multi-scale robust fusion (MRF) proposed by MRCNet enhances aggregation across semantic scales to address pose noise, while the motion enhancement mechanism (MEM) captures motion context to compensate for information blur caused by moving objects.

Furthermore, DiscoNet\cite{li2021learning} adopts a teacher-student framework from knowledge distillation, enhancing performance and bandwidth efficiency through early and mid-stage collaboration, significantly reducing model complexity. V2VNet\cite{wang2020v2vnet} proposes an advanced vehicle-to-vehicle (V2V) communication mechanism that utilizes spatially aware graph neural networks (GNNs) to integrate information from multiple autonomous vehicles, enabling intelligent information fusion across different viewpoints and time points.

\subsubsection{V2I Communication and Collaborative Perception}

In the field of V2I (Vehicle-to-Infrastructure) cooperation, while facing similar challenges to V2V (Vehicle-to-Vehicle) cooperation, several advanced methods and frameworks are driving the progress of this technology. Coop3D\cite{han2023collaborative} is a method that enhances detection accuracy and extends perception range by integrating the perception results of multiple agents. It provides an in-depth analysis of the advantages and disadvantages of early, mid, and late fusion, pointing out that although early fusion offers the best accuracy, its high demand for communication bandwidth could become a bottleneck in practical applications.

To address the challenges of heterogeneity and information sharing in V2I systems, V2X-ViT\cite{xu2022v2x} introduces the Vision Transformer (ViT) as its core architecture, combined with Heterogeneous Multi-Agent Self-Attention (HMSA) and Multi-Scale Window Self-Attention (MSwin) modules. These modules effectively tackle the challenges of heterogeneity, asynchronous information sharing, and localization errors in V2X systems. V2X-PC\cite{liu2024v2x} adopts a point cluster framework, integrating the structural and semantic information of objects, thereby overcoming the limitations of traditional dense BEV (Bird's Eye View) maps. Through point cluster packaging and aggregation modules, V2X-PC significantly improves scene perception accuracy while controlling bandwidth, and by employing innovative methods to address time delays and pose errors, it enhances the system's robustness in practical applications. CoopDet3D\cite{zimmer2024tumtraf} further validates the effectiveness of vehicle-road cooperative perception compared to other cooperative methods, and has introduced the TUMTraf dataset along with open-source annotation tools, providing a solid foundation for future research.

Transmission delay in vehicle-road cooperation is also an issue that cannot be ignored. An innovative method for vehicle-road cooperative 3D object detection has been proposed\cite{yu2023vehicle,yu2024flow}, which improves detection accuracy and efficiency through the prediction of feature flows. Traditional vehicle-road cooperative perception methods often face limitations in communication bandwidth and delays, but this study introduces a feature flow prediction model that enables efficient transmission and sharing of feature information between vehicles and infrastructure.

\subsubsection{Summary and Analysis}

Although collaborative perception technology shows great potential in autonomous driving and intelligent transportation systems by effectively addressing issues like single-vehicle blind spots and redundant computations, it still faces several technical challenges. These include verifying the authenticity of data, high implementation costs, and technical compatibility issues. Additionally, the current scale, quality, and richness of datasets limit the development of this work. Future research needs to address these issues to promote the widespread application and development of collaborative perception technology. Furthermore, with the advancement of technology and the emergence of new technologies such as 6G communication, it is expected that collaborative perception technology will be further advanced, enhancing road traffic safety and efficiency.
%%====================================================================================
\section{Discussion}
In the field of autonomous driving, the development of 3D perception technologies is moving towards more streamlined and comprehensive methods. These emerging technologies integrate data from various sensors, such as cameras, LiDAR, and radar, to provide a more complete and accurate environmental perception. Multimodal fusion methods, which combine data from different sensors, overcome the limitations of individual sensors. For instance, while LiDAR provides precise depth information, cameras capture rich color and texture details; the fusion of these data significantly enhances detection accuracy and robustness. Additionally, temporal perception technologies leverage information across consecutive frames, further improving the detection and tracking of objects in dynamic scenes and mitigating the risks associated with system latency. 3D occupancy grids allow for more granular spatial detection, and V2X technologies expand the detection range by connecting vehicles with other smart devices, thereby enhancing overall perception capabilities. End-to-end autonomous driving frameworks attempt to simplify system architecture and increase response speed by directly generating driving commands from raw sensor data through a unified network structure.

The continuous advancements in these technologies have led to significant performance improvements in autonomous driving systems, particularly in complex and dynamic environments. However, despite the promise of end-to-end frameworks as a future direction, their implementation often comes with high computational costs and complex model designs. For many research teams, training and deploying these large network models require substantial hardware support and extensive computational resources, which may not always be feasible in practical research and application contexts. In this regard, standalone 3D object detection technology still holds an important position. While it may not garner the same attention as 3D occupancy grids or end-to-end frameworks, 3D object detection does not represent outdated technology. On the contrary, it offers unique advantages in computational efficiency, resource demands, and broad applicability. 3D object detection methods can provide accurate and reliable detection results without relying on complex fusion techniques or enormous computational resources. This approach also demonstrates significant potential in fields beyond autonomous driving, such as robotic navigation, drone flight, AR/VR, and intelligent surveillance.

Looking ahead, as vehicle-to-everything (V2X) technology, end-to-end autonomous driving, and other advancements continue to develop and merge, autonomous driving systems will achieve higher levels of safety and intelligence. Real-time information exchange between vehicles, traffic infrastructure, other vehicles, and pedestrians will enable a more comprehensive understanding of the environment, allowing for faster and more accurate driving decisions. Moreover, with the development of 5G networks and future 6G technology, as well as the increasing computational capabilities of vehicles, real-time data transmission and processing will be significantly enhanced, supporting larger-scale cooperative perception and cloud computing. These advancements will not only drive the application of autonomous driving technology in complex scenarios but also lead smart transportation systems toward greater efficiency and safety.

\section{Conclusion}

This paper provides a comprehensive review of the current state and development trends of 3D object detection technologies, analyzing various methods based on cameras, LiDAR, and multisensor fusion. We have highlighted the strengths and weaknesses of each approach, along with their potential applications in autonomous driving. Additionally, we explored emerging trends in environmental perception technologies for autonomous driving, including temporal perception, 3D occupancy grids, end-to-end autonomous driving, and cooperative perception. By comparing different methods, this paper demonstrates their advantages in various scenarios and highlights the differences in hardware requirements. Researchers should select the most suitable approach based on their specific conditions rather than following trends blindly. This study provides a clear technological roadmap and methodological guidance for researchers in the field of autonomous driving, offering valuable insights for the development and optimization of future autonomous driving systems through an in-depth analysis of 3D object detection technologies.

%%====================================================================================
%% The Appendices part is started with the command \appendix;
%% appendix sections are then done as normal sections
%%====================================================================================
% \appendix
% \section{Example Appendix Section}
% \label{app1}

% Appendix text.

% %% For citations use: 
% %%       \cite{<label>} ==> [1]

% %%
% Example citation, See \cite{lamport94}.
%%====================================================================================
%% If you have bib database file and want bibtex to generate the
%% bibitems, please use
%%
%%  \bibliographystyle{elsarticle-num} 
%%  \bibliography{<your bibdatabase>}

%% else use the following coding to input the bibitems directly in the
%% TeX file.

%% Refer following link for more details about bibliography and citations.
%% https://en.wikibooks.org/wiki/LaTeX/Bibliography_Management

% \begin{thebibliography}{00}
% \bibliographystyle{elsarticle-num}
% \bibliography{refs}

\bibliographystyle{elsarticle-num}
\bibliography{refs}
\end{document}